\documentclass{article} 
\usepackage{iclr2026_conference,times}

\usepackage{amsmath,amsfonts,bm}

\def\eqref#1{equation~\ref{#1}}

\def\1{\bm{1}}

\DeclareMathAlphabet{\mathsfit}{\encodingdefault}{\sfdefault}{m}{sl}
\SetMathAlphabet{\mathsfit}{bold}{\encodingdefault}{\sfdefault}{bx}{n}

\usepackage{latexsym}
\usepackage{inconsolata}

\usepackage[T1]{fontenc}

\usepackage{microtype}

\usepackage{graphicx}
\usepackage{wrapfig}
\usepackage{enumitem}

\usepackage{booktabs}

\usepackage{amsmath}

\usepackage{pifont}
\usepackage{hyperref}       
\usepackage{url}            
\usepackage{booktabs}       
\usepackage{amsfonts}       
\usepackage{nicefrac}       
\usepackage{microtype}      
\usepackage{xcolor}         
\usepackage{textcomp}
\usepackage{graphicx}
\usepackage{multirow}
\usepackage{color}
\usepackage[ruled,vlined,linesnumbered]{algorithm2e}

\usepackage{threeparttable} 
\usepackage{tablefootnote} 
\usepackage{arydshln} 
\usepackage{amssymb} 
\usepackage{pifont} 
\usepackage{wrapfig}
\usepackage{verbatim}
\usepackage{subscript}
\usepackage{enumitem}
\usepackage{amsmath}
\usepackage{multirow}
\usepackage{colortbl}
\usepackage{booktabs}
\usepackage{hhline}
\usepackage{subcaption}
\usepackage{makecell}
\usepackage{mathrsfs}
\usepackage{tikz}
\usepackage{tabularx}
\usepackage{xstring}
\usepackage{ragged2e}
\usepackage[most]{tcolorbox}
\usepackage{hyperref}
\usepackage{xurl}
\usepackage[table,x11names]{xcolor}
\usepackage{booktabs}

\usepackage{xcolor}

\newcommand{\MYMETHOD}{\textsc{Mass}}

\title{\textcolor{cyan}
{\MYMETHOD}: \textcolor{cyan}{M}ulti-\textcolor{cyan}{A}gent \textcolor{cyan}{S}imulation \textcolor{cyan}{S}caling \\for Portfolio Construction}

\author{  \vspace{-25pt}\\
  \textbf{Taian Guo$^{1,3}$,\quad Haiyang Shen$^{2,1,\dag}$,\quad Jinsheng Huang$^{1,3}$,\quad Zhengyang Mao$^{1,3}$, Junyu Luo$^{1}$, BinQi Chen$^{1,3}$,}\\\textbf{Zhuoru Chen$^{3}$, \quad Xuhui Liu$^{3}$, \quad Bingyu Xia$^{3}$, \quad Luchen Liu$^{3,\dag}$, \quad Yun Ma$^{2,1,}$\textsuperscript{\ding{41}},\quad Ming Zhang$^{1,}$\textsuperscript{\ding{41}}}\vspace{8pt}\\
  $^{1}$School of Computer Science, Peking University \\
  $^{2}$Institute for Artificial Intelligence, Peking University \\
  $^{3}$Zhengren Quant, Beijing, China \\
  \texttt{\small \{taianguo, hyshen\}@stu.pku.edu.cn},\\
\texttt{\small \{mayun, mzhang\_cs\}@pku.edu.cn}, \vspace{8pt}\\
  Explore codes and datasets at:~\, \url{https://github.com/gta0804/MASS}\\
  \vspace{3em}
  $^\dag$Project Leader, \textsuperscript{\ding{41}}Corresponding Author
  \vspace{-48pt} \\
}

\iclrfinalcopy

\begin{document}

\maketitle

\begin{abstract}
The application of LLM-based agents in financial investment has shown significant promise, yet existing approaches often require intermediate steps like predicting individual stock movements or rely on predefined, static workflows. These limitations restrict their adaptability and effectiveness in constructing optimal portfolios. In this paper, we introduce the Multi-Agent Scaling Simulation (MASS), a novel framework that leverages multi-agent simulation for direct, end-to-end portfolio construction. At its core, MASS employs a backward optimization process to dynamically learn the optimal distribution of heterogeneous agents, enabling the system to adapt to evolving market regimes. A key finding enabled by our framework is the exploration of the scaling effect for portfolio construction: we demonstrate that as the number of agents increases exponentially (up to 512), the aggregated decisions yield progressively higher excess returns. Extensive experiments on a challenging, self-collected dataset from the 2023 Chinese A-share market show that MASS consistently outperforms seven state-of-the-art baselines. Further backtesting, stability analyses and the experiment on data leakage concerns validate its enhanced profitability and robustness. We have open-sourced our code, dataset, and training snapshots at \url{https://github.com/gta0804/MASS/} to foster further research.
\end{abstract}

\section{Introduction}
\label{sec:introduction}

The application of LLM-based agents in investment analysis has recently garnered significant attention from both academia and industry~\citep{tang2025alphaagentllmdrivenalphamining, TradingAgents2025, rdagentquant}. By assigning LLMs with distinct roles and providing them with relevant financial context, researchers have developed agent-based systems to tackle complex tasks such as alpha factor mining~\citep{chainofalpha, rdagentquant} and stock trend prediction~\citep{SEP2024, fincon2024}. These pioneering efforts highlight the potential of LLMs to process and reason over vast amounts of multi-modal financial data, including news, reports, and market indicators.

Despite their promise, existing LLM-based approaches in finance exhibit two primary limitations. First, many systems are designed for individual stock forecasting~\citep{SEP2024, fincon2024, TradingAgents2025}. While useful, predicting the movement of single stocks does not directly translate to constructing an optimal portfolio, which requires a holistic assessment of asset correlations, market sentiment, and risk diversification. Second, these systems typically rely on predefined procedural workflows to orchestrate agent interactions~\citep{ijcai2024p890}. This reliance on static, pre-programmed processes limits their ability to adapt to the highly dynamic and non-stationary nature of financial markets, potentially compromising their performance during market regime shifts.

In this paper, we introduce the Multi-Agent Scaling Simulation (\textbf{\MYMETHOD}) to address these challenges. \MYMETHOD~shifts the paradigm from individual stock prediction to direct portfolio construction by simulating a market of heterogeneous investor agents. Instead of relying on static workflows, \MYMETHOD~introduces a novel backward optimization process. This mechanism uses historical market data to dynamically learn the optimal distribution of agent types that maximizes portfolio returns, allowing the system to adapt its strategy. This approach provides \MYMETHOD~ with three key advantages: (1) It leverages aggregated information from a multi-agent simulation for direct, end-to-end portfolio construction, bypassing intermediate steps like individual stock prediction; (2) It replaces predefined workflows with a data-driven optimization process, enhancing adaptability and performance; and (3) It enables us to explore the multi-agent scaling effect for portfolio construction: as the number of agents increases exponentially, the system's decisions achieve higher excess returns. To the best of our knowledge, \MYMETHOD~is the first work to scale multi-agent simulation for this task up to 512 agents.

To rigorously evaluate \MYMETHOD, we collected a rather comprehensive and challenging dataset from the Chinese A-share market for the entirety of 2023, a period marked by significant volatility and two major market shifts. Our dataset, covering the \textit{SSE50}, \textit{CSI 300}, and \textit{ChiNext 100} indices, includes detailed firm-level features and macroeconomic indicators. Extensive experiments demonstrate that \MYMETHOD~significantly outperforms seven state-of-the-art baselines. Rigorous backtesting further confirms its ability to generate consistent excess returns with lower drawdowns. To address concerns about data leakage in the LLM (\texttt{Qwen-2.5-72B-Instruct}), we validate our findings on new data from the first quarter of 2025. Additional experiments confirm \MYMETHOD's scalability, stability in dynamic markets, and robustness to hyperparameters, while visualizations of the agent distribution dynamics offer insights into its adaptive mechanism.

In summary, this paper makes the following contributions:
\begin{itemize}[left=0.2cm]
\item We introduce \MYMETHOD, a novel framework leveraging multi-agent simulation with end-to-end backward optimization for decision-making in portfolio construction.
\item To our best knowledge, we are the first to explore and demonstrate a scaling effect in multi-agent simulation for portfolio construction, expanding the number of agents up to 512.
\item Extensive experiments show that \MYMETHOD~outperforms state-of-the-art baselines, delivering consistent excess returns, scalability, and stability. We also address potential data leakage concerns and validate our simulation's effectiveness through visualization.
\item We have introduced and released a comprehensive, realistic, and rich dataset, along with our code and training snapshots, to facilitate future research in this domain.
\end{itemize}
\section{Related Work}
\label{sec:RelatedWork}


This section reviews related work across three key areas to contextualize our research. We first discuss our primary research domain: existing investment analysis approaches within the financial market. We then survey the landscape of LLM-based multi-agent systems, which constitute our methodological approach. Finally, we cover the emerging research on scaling effects for multi-agent systems, a significant finding that informs our understanding of system performance.
\subsection{Investment Analysis}
\label{sec:investment_analysis}

Investment analysis research traditionally focuses on two main tracks: formulaic alpha mining  and stock price trend prediction. Alpha mining aims to discover mathematical expressions from financial data that predict future returns, using techniques like genetic algorithms~\citep{chen2021empirical}, deep reinforcement learning~\citep{alphagen, AlphaForge}, and more recently, LLM-based agents~\citep{tang2025alphaagentllmdrivenalphamining, chainofalpha, ding2025alphaeval, rdagentquant, shi2025navigating}. Stock price trend prediction employs methods ranging from traditional time-series analysis~\citep{choi2018stockpricecorrelationcoefficient} and deep learning models~\citep{2021DTML, xu2021hist, luo-etal-2023-causality,cikm2024_stock_explainable,NEURIPS2024_casualstock, umikdd2025, chen2025enhancer}. reinforcement learning models ~\citep{niu2022metatrader, yuanbilevel} to the latest LLM-based agents~\citep{SEP2024, TradingAgents2025, FinAgent2024} and foundation model training ~\citep{liu2025finr1largelanguagemodel, xiao2025tradingr1financialtradingllm, shi2025kronos}. While effective to a degree, alpha mining often treats the market monolithically, overlooking stock-specific idiosyncrasies, while most trend prediction methods focus on individual assets rather than portfolio-level optimization. Furthermore, many recent LLM-agent approaches rely on fixed, predefined workflows, limiting their adaptability. Additionally, LLMs trained on massive historical data introduce the risk of data leakage, as the historical data may encapsulate past market information.

\MYMETHOD~distinguishes itself from these works by shifting the focus from individual stock prediction or factor mining to the direct task of portfolio construction. Unlike methods that rely on predefined workflows, \MYMETHOD~employs a data-driven, end-to-end optimization framework to dynamically infer the underlying distribution of investor archetypes that leads to optimal portfolio performance. This simulation-based approach allows \MYMETHOD~to holistically model market dynamics and adapt to changing conditions, offering superior performance and market adaptability compared to forecasting individual asset movements in isolation.

\subsection{LLM-based Multi-Agent Systems}
\label{sec:llm_based_multi_agent}



LLM-based multi-agent systems (MAS) are broadly classified into two categories: \textit{Simulation} and \textit{Application}~\citep{ijcai2024p890}. Simulation-focused MAS are used to model emergent social~\citep{park2023generative}, economic~\citep{10.5555/3692070.3694596,li2023tradinggpt}, or psychological phenomena~\citep{kovac2023the,zhang-etal-2024-exploring}. Their primary goal is to validate existing theories or generate analytical insights. In contrast, Application-focused MAS employ specialized agents organized in structures like layers~\citep{liu2024dynamicllmpoweredagentnetwork} or centralized hierarchies~\citep{qian2025scaling} to collaboratively execute specific tasks, such as software development~\citep{10.5555/3666122.3668386} or scientific debate~\citep{10.5555/3692070.3692537}. These systems typically follow predefined procedural workflows to ensure efficient coordination.

\MYMETHOD~bridges the gap between these two categories. Compared to existing simulations, which are primarily used for analysis, \MYMETHOD~utilizes the aggregated output of its simulation for concrete, real-world decision-making, thereby expanding the practical boundaries of multi-agent simulation. Unlike existing applications that depend on rigid, predefined processes, \MYMETHOD~leverages a data-driven, end-to-end backward optimization mechanism. This allows the system to learn its own optimal collaborative strategy from market feedback, resulting in superior performance and adaptability without the need for hand-crafted workflows.

\subsection{Scaling Effects in Multi-Agent Systems}
\label{sec:scaling_law}

The study of scaling effects—predictable performance improvements with increased model size, data, or compute—is a key component of modern LLM research~\citep{scaling_law_openai}. One notable study explores cooperative scaling effects for various predefined agent architectures (e.g., linear, tree), expanding the agent count up to 64~\citep{qian2025scaling}. Another recent work ~\citep{dang2025multi} proposes an evolving orchestration where an RL-trained puppeteer dynamically organizes agents into cost-effective collaboration topologies, enhancing scaling in MAS.

As for scaling effects in MAS on the financial domain, Mars~\citep{2025mars} investigates the effect of training data scale on the realism of financial market simulations, but focuses on simulating the limited order books given by investors rather than the investors within it. StockAgent ~\citep{zhang2024aimeetsfinancestockagent} introduces an LLM-driven multi-agent system to simulate investor trading behaviors in an LLM-generated stock market environment in response to external factors and market dynamics. TwinMarket ~\citep{twinmarket} is a novel scalable multi-agent framework leveraging Large Language Models to simulate investor behavior in financial markets through dynamic social networks. These studies either constrain agent interactions to fixed topologies or do not focus on utilizing the simulation results to guide real market investments.

In contrast, \MYMETHOD~introduces and investigates a novel scaling paradigm for multi-agent decision-making. Our scaling effect does not rely on a prescribed form of cooperation. Instead, each agent is given a partial view of the market, and as the number of agents increases, the system's collective awareness of the market grows. The core challenge, which we address via our backward optimization process, is learning how to aggregate this distributed intelligence to achieve a specific real-world objective (i.e., maximizing portfolio returns). \MYMETHOD~is the first work to demonstrate this scaling effect in a financial application, expanding the number of simulated agents to 512 and showing a clear correlation between agent scale and investment performance.

\section{Method}
\label{sec:method}

\begin{figure}[h!]
\centering
\includegraphics[width=\linewidth]{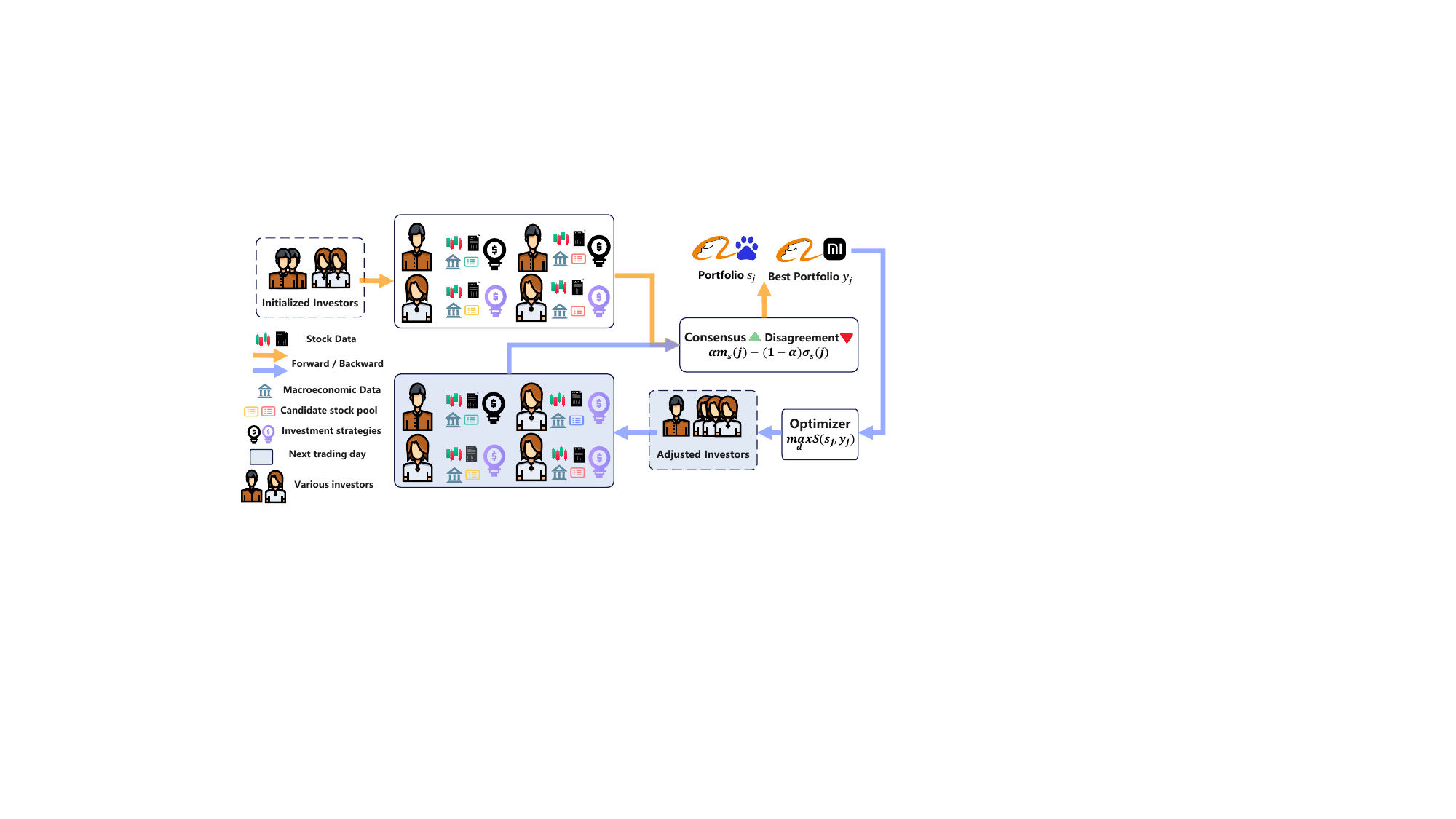}
\caption{\MYMETHOD~ operates in a loop consisting of forward propagation and backward optimization. In the forward propagation, \MYMETHOD~ initializes investors using the previous day's investor distribution along with today's stock and macroeconomic data. It then constructs portfolios $s_j$ based on the market disagreement hypothesis. During backward optimization, an optimizer updates the investor distribution, which is then passed to the next trading day.}
\label{sec::method::figure}
\end{figure}

In this section, we introduce \MYMETHOD, a novel framework that formulates portfolio construction as a dynamic online learning problem. The core idea is to simulate a market of heterogeneous investor agents and learn to optimally aggregate their diverse decisions. \MYMETHOD~ operates in a daily cycle of two key processes: \textbf{Forward Propagation}, where agents generate investment signals for the current day, and \textbf{Backward Optimization}, which refines the model by learning from the previous day's outcomes. This adaptive loop, illustrated in Figure \ref{sec::method::figure}, allows \MYMETHOD~ to continuously adjust to evolving market conditions. The overall procedure is formalized in Algorithm \ref{sec::Appendix::code}.

\subsection{Forward Propagation}

The forward propagation process simulates market activity on a given day $j$ to produce a signal that guides the construction of the portfolio. This involves initializing a diverse population of agents, executing their investment strategies, and aggregating their collective decisions based on a multi-modal stock dataset $\mathcal{X}$.

\subsubsection{Investor Initialization}
\label{sec::method::investor_initialization}
To capture the diverse perspectives within a real market, \MYMETHOD~initializes a population of $N = n^{\text{type}} \times n^{\text{inv}}$ agents. These agents are categorized into $n^{\text{type}}$ distinct types, each embodying a unique investment style (e.g., style outline, risk appetite, rationality). This heterogeneity is crucial for creating a rich and realistic simulation. Each agent type $i$ is provided access to a specific subset of multi-modal data $\mathcal{X}_i \subset \mathcal{X}$. Furthermore, to model the practical constraint that no single investor can monitor the entire market, each individual agent $(i, k)$ is assigned a static, random subset of stocks, denoted as $\text{Pool}(i, k)$, where $|\text{Pool}(i, k)| = n^{\text{sel}}$. This design choice also manages the context length limitations of the underlying LLM. The design details of investor initialization are in Appendix \ref{sec::Appendix:promptInitalization}.

\subsubsection{Investment Strategy Execution}
On each trading day $j$, agents first formulate a daily strategy and then make investment decisions.
First, to ensure strategies are adaptive to the prevailing economic climate, each agent type $i$ generates a daily investment strategy by interpreting the latest macroeconomic data $\mathbf{M}_j$ within the context of its intrinsic style. This is performed by an LLM-based function $F_1$:
\begin{equation}
\label{sec::methods::strgetgy_initalization}
    \text{Strategy}_{i,j}  = F_1\bigl(\text{StyleDesc}_i, \mathbf{M}_j \bigr)
\end{equation}
where $\text{StyleDesc}_i$ is the textual description of agent type $i$'s investment philosophy.

Next, each agent $(i, k)$ applies this daily strategy to its observable stock universe $\text{Pool}(i, k)$. The agent analyzes the relevant features for these stocks and selects a subset for investment. This decision is modeled by a second LLM-based function $F_2$:
\begin{equation}
\label{sec::methods::investor_decision}
\begin{split}  
    \text{Codes}_{i,k,j} = F_2\Bigl(  
        \text{Strategy}_{i,j}, \; 
        \{\text{data for } s \in \text{Pool}(i, k)\}, \;
        \text{StyleDesc}_i
    \Bigr)
\end{split} 
\end{equation}
where $\text{Codes}_{i,k,j} \subseteq \text{Pool}(i, k)$ is the set of stocks selected by agent $(i, k)$ on day $j$. The design details of this section are provided in Appendix \ref{sec::Appendix:promptExecution}.

\subsubsection{Score Aggregation}
To derive an actionable signal for each stock, we aggregate the decisions from all $N$ agents. Our aggregation strategy is grounded in the market disagreement hypothesis~\citep{miller1977risk,diether2002differences}, which posits that stocks with high consensus and low disagreement among investors tend to yield higher future returns. This provides a theoretically sound basis for combining agent outputs. We provide more details about market disagreement hypothesis on Appendix ~\ref{sec::Appendix::marketdis}.

Let $V_{i,s,j}$ be the fraction of agents of type $i$ that selected stock $s$ on day $j$. Let $\mathbf{d}_{j-1} = [d_{1, j-1}, \dots, d_{n^{\text{type}}, j-1}]^\top$ be the distribution of agent types, optimized from the previous day. We quantify consensus and disagreement for each stock $s$ by computing the weighted mean ($m_s$) and weighted standard deviation ($\sigma_s$) of selections across all agent types:

\begin{subequations} \label{sec::methods::score_agg_1}
\begin{align}
    m_s(j) &= \sum_{i=1}^{n^\text{type}} d_{i, j-1} \cdot V_{i,s,j} \quad \text{(Consensus)} \label{subeq:consensus} \\
    \sigma_s(j) &= \sqrt{\sum_{i=1}^{n^\text{type}} d_{i, j-1} (V_{i,s,j} - m_s(j))^2} \quad \text{(Disagreement)} \label{subeq:disagreement}
\end{align}
\end{subequations}

The final signal for each stock integrates these two components, rewarding consensus and penalizing disagreement:
\begin{equation}
\label{sec::methods::score_agg_2}
    \mathrm{Signal}(s,j) = \alpha \cdot m_s(j) - (1 - \alpha) \cdot \sigma_s(j)
\end{equation}
where $\alpha \in [0,1]$ is a hyperparameter balancing the two effects. This signal is then used to rank stocks and construct the daily portfolio $\mathbf{P}_j$.

\subsection{Backward Optimization}

A key innovation of \MYMETHOD~ is its ability to adapt to changing market regimes. This is achieved through the backward optimization process, which dynamically adjusts the agent type distribution $\mathbf{d}_j$ at the end of each day $j$. The objective is to find the distribution that would have yielded the best performance over a recent historical window, ensuring the model continuously learns from market feedback.

Specifically, at the end of day $j$, we define a look-back window\footnote{To avoid inadvertent use of future information, $\mathbf{Y}^{k}[:,j - k]$ is excluded, because this label depends on the first 15 minutes on day $j + 1$ and latency in live trading systems should be considered.} of size $\omega_{\text{opt}}$. We use the agent decisions $\{V_{i,s,t}\}$ and the actual market returns $\{\mathbf{Y}_t\}$ for the period $t \in [j-\omega_{\text{opt}}+1, j]$. For any candidate distribution $\mathbf{d}$, we can compute the historical signals $\mathrm{Signal}_{\mathbf{d}}(s,t)$ for this period. The goal is to find the optimal distribution $\mathbf{d}_j$ that maximizes the correlation between these historical signals and the actual returns. This is formulated as an optimization problem:
\begin{equation}
\label{sec::methods::distribution_optimization}
    \mathbf{d}_{j} = \arg\max_{\mathbf{d} \in \Delta^{n^{\text{type}}-1}} \; \mathcal{S} \left( \{\mathrm{Signal}_{\mathbf{d}}(:,t)\}_{t=j-\omega_{\text{opt}}+1}^{j}, \; \{\mathbf{Y}_t\}_{t=j-\omega_{\text{opt}}+1}^{j} \right)
\end{equation}
where $\Delta^{n^{\text{type}}-1}$ is the probability simplex, and $\mathcal{S}$ is a similarity metric such as the Rank Information Coefficient (RIC). We employ simulated annealing~\citep{kirkpatrick1983optimization} as the optimizer $\mathcal{O}$ to solve this problem. The resulting distribution $\mathbf{d}_j$ is then carried forward to the next day's forward propagation step (Eq. \ref{sec::methods::score_agg_1}), completing the online learning cycle.

\section{Evaluation}
\label{sec:Evaluation}

\begin{table*}[htbp]
\centering
\caption{Comparisons with baselines and the experiment on data leakage concern. \MYMETHOD~outperforms all others across all 3 stock pools. The best performance in each column is highlighted in \textbf{bold}. For more evaluation metrics on portfolio construction, please refer to Appendix ~\ref{sec::Appendix::furtherRes}.}
\label{tab:1}
\scalebox{0.72}{
\begin{tabular}{@{}lrrrrrrrrrrrr@{}}
\toprule
\multicolumn{13}{c}{\textbf{Main Experiments (Throughout 2023)}} \\
\midrule
\multicolumn{1}{c}{\multirow{2}{*}{\textbf{Method}}} & \multicolumn{4}{c}{\textbf{SSE50}} & \multicolumn{4}{c}{\textbf{CSI 300}} & \multicolumn{4}{c}{\textbf{Chi Next 100}} \\
\cmidrule(lr){2-5} \cmidrule(lr){6-9} \cmidrule(lr){10-13}
\multicolumn{1}{c}{} & \multicolumn{1}{c}{RIC} & \multicolumn{1}{c}{RICIR} & \multicolumn{1}{c}{IC} & \multicolumn{1}{c}{ICIR} & \multicolumn{1}{c}{RIC} & \multicolumn{1}{c}{RICIR} & \multicolumn{1}{c}{IC} & \multicolumn{1}{c}{ICIR} & \multicolumn{1}{c}{RIC} & \multicolumn{1}{c}{RICIR} & \multicolumn{1}{c}{IC} & \multicolumn{1}{c}{ICIR} \\
\midrule
Proxy Indicator~\citep{diether2002differences} & 3.82 & 19.73 & 2.89 & 16.63 & 3.84 & 30.44 & 3.60 & 27.03 & -0.94 & -7.05 & 0.16 & 1.29 \\
LightGBM ~\citep{2017LGBM}    & 3.25 & 21.78 & 4.51 & 27.30 & 5.20 & 36.06 & 3.19 & 23.62 & 2.94 & 30.69 & 0.88 & 8.70 \\
DTML~\citep{2021DTML}         & 5.04 & 28.15 & 4.93 & 26.71 & 4.91 & 35.72 & 4.17 & 31.10 & 3.45 & 26.55 & 3.21 & 21.97 \\
MASTER~\citep{li2024master}   & 5.13 & 28.37 & 4.97 & 27.01 & 5.01 & 35.47 & 4.23 & 30.78 & 3.92 & 31.03 & 4.07 & 28.62 \\
SEP~\citep{SEP2024}           & 4.79 & 27.56 & 4.16 & 26.40 & 3.83 & 5.42  & 0.61 & 7.65  & 4.81 & 34.88 & 5.29 & 36.98 \\
FinCON~\citep{fincon2024}     & 4.88 & 26.18 & 4.35 & 25.67 & 0.70 & 9.57  & 0.96 & 13.42 & 5.01 & 37.18 & 5.53 & 40.54 \\
TradingAgents~\citep{TradingAgents2025} & 4.92 & 27.71 & 4.33 & 25.69 & 3.01 & 10.14 & 1.02 & 14.80 & 5.37 & 38.15 & 5.60 & 41.06 \\
\midrule
\rowcolor{gray!20}
\textbf{\MYMETHOD} & \textbf{{8.16}} & \textbf{{41.74}} & \textbf{{5.90}} & \textbf{{33.43}} & \textbf{{6.50}} & \textbf{{43.49}} & \textbf{{4.65}} & \textbf{{33.32}} & \textbf{{7.62}} & \textbf{62.87} & \textbf{{6.28}} & \textbf{{55.88}} \\
\midrule
\multicolumn{13}{c}{\textbf{Experiments on data leakage concern (The first quarter of 2025)}} \\
\midrule
\multicolumn{1}{c}{\multirow{2}{*}{\textbf{Method}}} & \multicolumn{4}{c}{\textbf{SSE50}} & \multicolumn{4}{c}{\textbf{CSI 300}} & \multicolumn{4}{c}{\textbf{CSI A500}} \\
\cmidrule(lr){2-5} \cmidrule(lr){6-9} \cmidrule(lr){10-13}
\multicolumn{1}{c}{} & \multicolumn{1}{c}{RIC} & \multicolumn{1}{c}{RICIR} & \multicolumn{1}{c}{IC} & \multicolumn{1}{c}{ICIR} & \multicolumn{1}{c}{RIC} & \multicolumn{1}{c}{RICIR} & \multicolumn{1}{c}{IC} & \multicolumn{1}{c}{ICIR} & \multicolumn{1}{c}{RIC} & \multicolumn{1}{c}{RICIR} & \multicolumn{1}{c}{IC} & \multicolumn{1}{c}{ICIR} \\
\midrule
\rowcolor{gray!20}
\textbf{\MYMETHOD} & 4.50 & 24.41 & 6.12 & 38.33 & 3.91 & 37.44 & 3.36 & 34.56 & 5.19 & 56.17 & 4.66 & 48.82 \\
\bottomrule
\end{tabular}
}
\end{table*}

\textbf{Optimization Strategy:} We employ simulated annealing~\citep{kirkpatrick1983optimization} as the optimizer in our backward optimization process to align the investor distribution with optimal market portfolios. 

\textbf{Complexity:} Although the total time complexity for a historical simulation is $O(n_{\text{type}} \times n_{\text{inv}} \times T)$, in a live trading scenario, the daily cost is only $O(n_{\text{type}} \times n_{\text{inv}})$. This is because we can store the latest agent distribution snapshot and update it with the newly arriving data stream. To ensure \MYMETHOD's reproductivity, a detailed analysis of time and computational costs is provided in Appendix \ref{sec::Appendix:moreDetail:timeandcost}.

\textbf{Dataset and Stock Pools:} While prior studies~\cite{SEP2024, FinAgent2024, TradingAgents2025} provide valuable insights, their evaluations often focus on US markets during stable bull periods~\cite{Nasdaq100}. To test model robustness in a more volatile context, and lacking a comparable multimodal US dataset, we introduce a new dataset from the Chinese A-share market. Our dataset covers the entirety of 2023, a period marked by high volatility and two major shifts, thus offering a challenging benchmark. To foster further research, we have open-sourced one of our dataset. The data covers three key indices: \textit{SSE 50}~\citep{SSE50_methodology}, \textit{CSI 300}~\citep{CSI300_methodology}, and \textit{ChiNext 100}~\citep{ChiNext100_methodology}. For each stock, the dataset includes news, financial reports, price-volume features, and fundamental data, complemented by macroeconomic indicators. Details about the construction of our dataset are in Appendix~\ref{sec:Appendix:dataset}.

\textbf{Baselines:} We compare \MYMETHOD~ with various baselines across different categories: a traditional proxy indicator~\citep{diether2002differences}; a machine learning model, LightGBM~\citep{2017LGBM}; deep learning models, DTML~\citep{2021DTML} and MASTER ~\citep{li2024master}; and three SOTA agent-based methods, SEP~\citep{SEP2024}, FINCON~\citep{fincon2024}, and TradingAgents~\citep{TradingAgents2025}. While Mars~\citep{2025mars} is relevant, a direct comparison is not possible because their model weights are still under review ~\footnote{~\url{https://github.com/microsoft/MarS}}. Further details about baseline descriptions and our implementations are in Appendix~\ref{sec:Appendix:baselines}.

\textbf{Metrics:} We use four standard metrics to assess both correlation and consistency: the Information Coefficient (IC) and Rank Information Coefficient (RIC) quantify Pearson and Spearman correlations between predicted ($Signal$) and actual returns ($r$), respectively. Their stability is measured by the Information Coefficient Information Ratio (ICIR) and Rank Information Coefficient Information Ratio (RICIR), defined as $\mathbb{E}[\text{IC}] / \text{Std}(\text{IC})$ and $\mathbb{E}[\text{RIC}] / \text{Std}(\text{RIC})$. Besides, to ensure the robustness of our evaluation process, we incorporate more metrics on Appendix ~\ref{sec::Appendix::furtherRes}.

\textbf{Experiment-Specific Settings:}
We utilize \textit{Qwen2.5 72B Instruct} ~\citep{qwen2025qwen25technicalreport} to implement \MYMETHOD. For the Main Experiments (Table \ref{tab:1}), we set $n^{\text{type}}=16$ and $n^{\text{inv}}=32$. For SSE 50 and ChiNext 100, $n^{\text{sel}}$ is 20; for the larger CSI 300, it is 30. We set $\alpha=0.5$ for SSE 50 and CSI 300. For ChiNext 100, we use $\alpha=0.2$, reflecting that for China's growth market, where valuations are often disconnected from fundamentals, disagreement factors ($\sigma_s$) are more predictive than consensus factors ($m_s$).For the backward optimization process, we employ simulated annealing (SA)~\citep{kirkpatrick1983optimization}. The key hyperparameters are configured as follows: an initial temperature of 40, a maximum of 100 iterations, and a cooling rate of 0.95. The optimizer look-back window $\omega_{opt}$ is set to 5. For the Data Leakage Experiments (Table \ref{tab:1}), we used \texttt{Qwen2.5-72B-Instruct} (released Sept. 2024) and evaluated it on data from Q1 2025 across SSE 50, CSI 300, and the new CSI A500 index~\citep{csia500_methodology}. For other experiments, we primarily used the CSI 300 pool due to its size and popularity, unless specified otherwise.

\subsection{Results and Analysis}
\label{sec:ResultsandAnalysis}

\subsubsection{Main Experiments}
\label{sec:MainExperiments}

\setlength{\tabcolsep}{3pt}

Table \ref{tab:1} presents the primary comparison against baselines. The key observations are twofold. First, \MYMETHOD~achieves the best performance across all metrics and stock pools, consistently outperforming the next-best methods (TradingAgents, SEP, FinCON and MASTER). Second, we observe that while agent-based methods like SEP and FINCON perform reasonably on smaller pools, their effectiveness diminishes significantly on the larger CSI 300. Our analysis indicates this is because their self-reflection mechanisms, which require processing extensive historical results in-context, face comprehension and decision-making challenges with an increasing number of stocks. \MYMETHOD~avoids this bottleneck as its architecture does not require any single agent to process vast global information, demonstrating superior scalability.


\subsubsection{Experiments on data leakage concern}
\label{sec:FurtherExperimentsUsingNewdataandStockPool}

To confirm that \MYMETHOD's performance is not merely a result of the LLM memorizing 2023 market data, we conducted rigorous tests on unseen data. As shown in the lower part of Table \ref{tab:1}, \MYMETHOD~maintains significant effectiveness on both unseen data from existing indices (SSE 50, CSI 300 in Q1 2025) and on a completely new stock pool (CSI A500). This result provides strong evidence that the model's success stems from its methodological framework rather than prior knowledge.


\subsubsection{Backtesting Experiments}
\label{sec:BacktestingExperiments}

\begin{figure}[h] 
\centering
\includegraphics[width=0.95\textwidth]{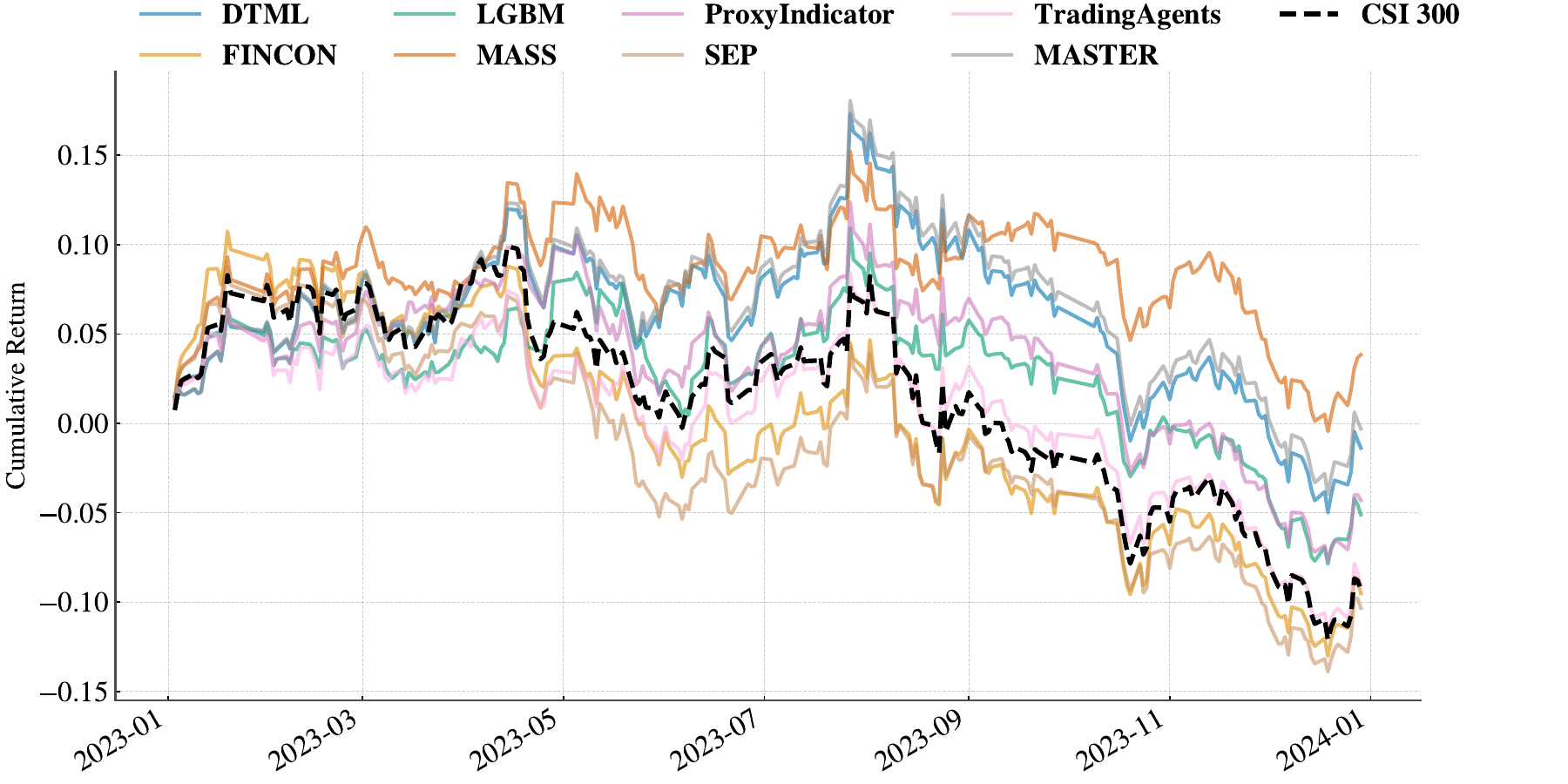}
\caption{Backtesting on the CSI 300 Stock Pool compared with Baselines and the CSI 300 Index.}
\label{fig:BacktestingExperiments}
\end{figure}

Figure \ref{fig:BacktestingExperiments} translates the statistical metrics into a practical financial outcome via backtesting. The plot of cumulative excess returns shows that \MYMETHOD~not only generates substantially higher returns than the baselines and the CSI 300 index but also maintains significantly lower drawdowns. This result highlights \MYMETHOD's dual advantages in both profitability and risk control, underscoring its real-world applicability. Backtesting implementation details are in Appendix \ref{sec::Appendix::backtesting}.


\subsubsection{Scaling Experiments}
\label{sec:ScalingExperiments}

\begin{figure*}[!ht]
    \centering
        \begin{minipage}[t]{0.48\textwidth}
            \centering
            \includegraphics[width=\linewidth]{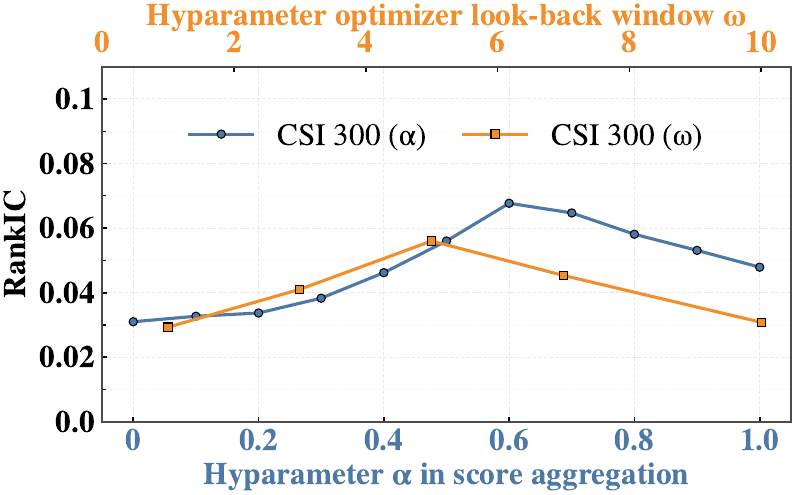}
            \caption{\MYMETHOD~exhibits a moderate sensitivity to changes in hyperparameters.}
            \label{fig:ParameterSensitivityExperiments}
        \end{minipage}
        \hfill 
        \begin{minipage}[t]{0.48\textwidth}
            \centering
            \includegraphics[width=\linewidth]{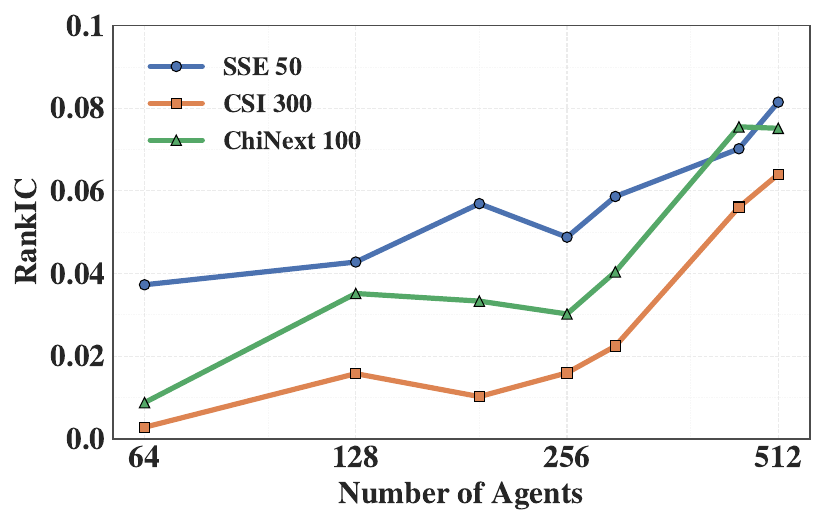}
            \caption{As the number of agents increases exponentially, \MYMETHOD~is able to obtain even more refined market information.}
            \label{fig:scaling_law}
        \end{minipage}
\end{figure*}


To verify the multi-agent scaling effect, we investigated the performance of \MYMETHOD~as we exponentially increased the number of agents ($n^{\text{type}} \times n^{\text{inv}}$) while keeping other parameters fixed. The results in Figure \ref{fig:scaling_law} show a clear, approximately linear growth in the RankIC metric as the total number of agents increases. This confirms that by simulating more agents, \MYMETHOD~is able to capture more refined market information, leading to better investment decisions. To the best of our knowledge, we are the first to explore this scaling effect in multi-agent simulation for portfolio construction, expanding the agent count up to 512.

\subsubsection{Ablation Studies}
\label{sec:ablation_studies}

\begin{table*}[ht!]
\centering
\caption{Ablation study results for CSP, PMD, BO, MDH, and an investigation of \MYMETHOD~, which daily updates the candidate stock pool, called \MYMETHOD(DU). The best performance is indicated in \textbf{bold}. The EMCL refers to the inability to operate when exceeding the maximum context length of the LLM.}
\label{tab:ablation_study}
\begin{tabular}{@{}lrrrrrrrrrrrr@{}}
\toprule
\multicolumn{1}{c}{\multirow{2}{*}{\textbf{Method}}} & \multicolumn{4}{c}{\textbf{SSE 50}} & \multicolumn{4}{c}{\textbf{CSI 300}} & \multicolumn{4}{c}{\textbf{Chi Next 100}} \\ 
\cmidrule(lr){2-5} \cmidrule(lr){6-9} \cmidrule(lr){10-13}
\multicolumn{1}{c}{} & \multicolumn{1}{c}{RIC} & \multicolumn{1}{c}{RICIR} & \multicolumn{1}{c}{IC} & \multicolumn{1}{c}{ICIR} & \multicolumn{1}{c}{RIC} & \multicolumn{1}{c}{RICIR} & \multicolumn{1}{c}{IC} & \multicolumn{1}{c}{ICIR} & \multicolumn{1}{c}{RIC} & \multicolumn{1}{c}{RICIR} & \multicolumn{1}{c}{IC} & \multicolumn{1}{c}{ICIR} \\ \midrule
w/o CSP & 1.65 & 11.19 & 1.67 & 11.73 & \multicolumn{4}{c}{EMCL} & \multicolumn{4}{c}{EMCL} \\
w/o PMD & 5.25 & 29.75 & 3.43 & 21.10 & 2.57 & 33.38 & 2.23 & 30.64 & 2.26 & 17.16 & 2.99 & 22.70 \\
w/o BO & 0.76 & 4.75 & -0.13 & -8.44 & 0.36 & 5.36 & 0.41 & 6.69 & 2.88 & 19.43 & 3.12 & 22.03 \\
w/o MDH & 6.28 & 32.68 & 3.85 & 25.39 & 4.65 & 31.03 & 2.98 & 27.86 & -3.12 & -28.93 & -2.46 & -26.44 \\ \midrule
\rowcolor{gray!20}
\textbf{\MYMETHOD(DU)} & 8.03 & 41.68 & 5.79 & \textbf{33.52} & 6.48 & 42.86 & 4.52 & 32.95 & \textbf{7.65} & \textbf{63.02} & \textbf{6.29} & \textbf{55.91} \\
\rowcolor{gray!20}
\textbf{\MYMETHOD} & \textbf{8.16} & \textbf{41.74} & \textbf{5.90} & 33.43 & \textbf{6.50} & \textbf{43.49} & \textbf{4.65} & \textbf{33.32} & 7.62 & 62.87 & 6.28 & 55.88 \\ \bottomrule
\end{tabular}
\end{table*}
\vspace{-1em}

Table \ref{tab:ablation_study} presents the results of ablating four key design choices and our variant of our proposed \MYMETHOD:
\begin{itemize}[left=0.2cm]
\item \textbf{w/o CSP (Candidate Stock Pool):} Removing this component causes the model to fail on larger indices due to exceeding the LLM's context length (EMCL). This confirms that CSP is essential for the system's scalability.
\item \textbf{w/o PMD (Provide Macro Data):} Removing macroeconomic data leads to a significant performance drop, as agents lack the context to make diverse, timely decisions, thus reducing system randomness and adaptability.
\item \textbf{w/o BO (Backward Optimization):} This is the most critical ablation study. Disabling the optimization process in Equation ~\ref{sec::methods::distribution_optimization} causes performance to collapse, yielding near-zero or negative IC values. This proves that the end-to-end, adaptive learning of agent distribution is the core mechanism driving \MYMETHOD's success.
\item \textbf{w/o MDH (Market Disagreement Hypothesis):} Relying solely on consensus led to a major performance drop and was even counterproductive on the ChiNext index, demonstrating the importance of our theory-grounded aggregation method.

\item \textbf{\MYMETHOD(DU) (Daily Updated Candidate Stock Pool)}:In Section ~\ref{sec::method::investor_initialization}, we construct each agents' a static candidate stock pool. To confirm the robustness of \MYMETHOD~ and eliminate the possible impact of this pre-defined set, we also test a variant which updates each agent's candidate stock pool on each trading day, finding its impact negligible. This suggests that the key is the partitioned view, not whether the view is static or dynamic.
\end{itemize}

\setlength{\tabcolsep}{5.0pt}

\subsubsection{Stability and Agent Distribution Visualization Experiment}
\label{sec:StabilityandAgentDistributionVisualizationExperiment}

\begin{figure}[ht]
\centering
\includegraphics[width=\linewidth]{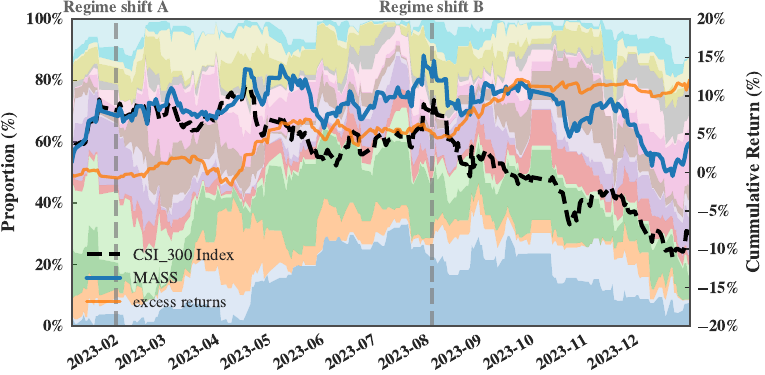}
\caption{The distribution of agents in \MYMETHOD~ swiftly adapts to changes in market styles (A and B), allowing it to consistently achieve stable excess returns compared to the CSI 300 Index.}
\label{fig:StabilityandAgentDistributionVisualizationExperiment}
\end{figure}

Figure \ref{fig:StabilityandAgentDistributionVisualizationExperiment} provides a visual proof of \MYMETHOD's adaptability. The background color tracks the temporal evolution of the agent distribution throughout 2023, with the left y-axis representing the proportion of different agent types. The right y-axis indicates the cumulative return, which is used to plot the performance of our  \MYMETHOD~ (blue line) and the CSI 300 Index (black dashed line). The orange line illustrates the excess cumulative return of \MYMETHOD~ compared to the CSI 300 benchmark. The two major market shifts in February (rebound to consolidation) and August (consolidation to decline) are marked as A and B. It is evident that during both transitions, \MYMETHOD's backward optimization mechanism swiftly adapted the agent distribution to align with the new market style. This rapid adaptation enabled \MYMETHOD~to consistently achieve stable excess returns compared to the CSI 300 index, even during periods of high volatility.

\subsubsection{Parameter Sensitivity Experiments}
\label{sec:ParameterSensitivityExperiments}

To investigate the sensitivity of \MYMETHOD~to its parameters, we analyzed two hyperparameters: the score aggregation weight $\alpha$ (Equation~\ref{sec::methods::score_agg_2}) and the optimizer look-back window $\omega_{\text{opt}}$. The parameter $\alpha$ manages the balance between disagreement and consensus components in portfolio construction, while $\omega_{\text{opt}}$ influences information capacity—too short a window limits it, whereas too long a duration hinders regime adaptation. The experimental results are presented in Figure~\ref{fig:ParameterSensitivityExperiments}.

We observe that although adjustments to these two hyperparameters lead to slight variations in system performance, these changes are within acceptable limits. This indicates that \MYMETHOD~exhibits a moderate sensitivity to parameter changes.
\section{Conclusion}
\label{sec:Conclusion}

In this paper, we introduce \MYMETHOD, a multi-agent scaling simulation framework designed for portfolio construction. \MYMETHOD~leverages large-scale agent simulations and a backward optimization process to achieve a comprehensive understanding of market dynamics. This approach offers various advantages, including enhanced scalability, robustness, and the ability to generate stable excess returns. 

In the future, we anticipate that the paradigm established by \MYMETHOD~ will extend beyond investment portfolio management to encompass a wider range of tasks, such as supply chain optimization, agricultural decision-making, and weather prediction.

\bibliography{custom}

\begin{thebibliography}{54}
\providecommand{\natexlab}[1]{#1}
\providecommand{\url}[1]{\texttt{#1}}
\expandafter\ifx\csname urlstyle\endcsname\relax
  \providecommand{\doi}[1]{doi: #1}\else
  \providecommand{\doi}{doi: \begingroup \urlstyle{rm}\Url}\fi

\bibitem[Bali et~al.(2023)Bali, Kelly, Mörke, and Rahman]{NBER2023}
Turan~G Bali, Bryan~T Kelly, Mathis Mörke, and Jamil Rahman.
\newblock Machine forecast disagreement.
\newblock Working Paper 31583, National Bureau of Economic Research, August 2023.
\newblock URL \url{http://www.nber.org/papers/w31583}.

\bibitem[Cao et~al.(2025)Cao, Xi, Liao, Yang, and Cao]{chainofalpha}
Lang Cao, Zekun Xi, Long Liao, Ziwei Yang, and Zheng Cao.
\newblock Chain-of-alpha: Unleashing the power of large language models for alpha mining in quantitative trading.
\newblock \emph{arXiv preprint arXiv:2508.06312}, 2025.

\bibitem[Chen et~al.(2021)Chen, Chen, and Du]{chen2021empirical}
Tianxiang Chen, Wei Chen, and Luyao Du.
\newblock An empirical study of financial factor mining based on gene expression programming.
\newblock In \emph{2021 4th International Conference on Advanced Electronic Materials, Computers and Software Engineering (AEMCSE)}, pp.\  1113--1117. IEEE, 2021.

\bibitem[Chen et~al.(2025)Chen, Li, Wang, and Wang]{chen2025enhancer}
Weijun Chen, Shun Li, Heyuan Wang, and Tengjiao Wang.
\newblock Enhancer: A distribution-aware framework with temporal-relational meta-learning for stock prediction.
\newblock In \emph{Proceedings of the 31st ACM SIGKDD Conference on Knowledge Discovery and Data Mining V. 2}, pp.\  250--261, 2025.

\bibitem[China Securities Index~Co.(2020)]{SSE50_methodology}
Ltd China Securities Index~Co.
\newblock \emph{Compilation of the SSE 50 Index}, 2020.
\newblock URL \url{https://oss-ch.csindex.com.cn/static/html/csindex/public/uploads/indices/detail/files/zh_CN/000016_Index_Methodology_cn.pdf}.

\bibitem[China Securities Index~Co.(2023)]{CSI300_methodology}
Ltd China Securities Index~Co.
\newblock \emph{Compilation of the CSI 300 Index}, 2023.
\newblock URL \url{https://oss-ch.csindex.com.cn/static/html/csindex/public/uploads/indices/detail/files/zh_CN/000300_Index_Methodology_cn.pdf}.

\bibitem[China Securities Index~Co.(2024)]{csia500_methodology}
Ltd China Securities Index~Co.
\newblock \emph{Compilation of the CSIA500 Index}, 2024.
\newblock URL \url{https://oss-ch.csindex.com.cn/static/html/csindex/public/uploads/indices/detail/files/zh_CN/000510_Index_Methodology_cn.pdf}.

\bibitem[Choi(2018)]{choi2018stockpricecorrelationcoefficient}
Hyeong~Kyu Choi.
\newblock Stock price correlation coefficient prediction with arima-lstm hybrid model, 2018.
\newblock URL \url{https://arxiv.org/abs/1808.01560}.

\bibitem[Dang et~al.(2025)Dang, Qian, Luo, Fan, Xie, Shi, Chen, Yang, Che, Tian, et~al.]{dang2025multi}
Yufan Dang, Chen Qian, Xueheng Luo, Jingru Fan, Zihao Xie, Ruijie Shi, Weize Chen, Cheng Yang, Xiaoyin Che, Ye~Tian, et~al.
\newblock Multi-agent collaboration via evolving orchestration.
\newblock In \emph{The Thirty-ninth Annual Conference on Neural Information Processing Systems}, 2025.

\bibitem[Diether et~al.(2002)Diether, Malloy, and Scherbina]{diether2002differences}
Karl~B Diether, Christopher~J Malloy, and Anna Scherbina.
\newblock Differences of opinion and the cross section of stock returns.
\newblock \emph{The Journal of Finance}, 57\penalty0 (5):\penalty0 2113--2141, 2002.
\newblock \doi{10.1111/0022-1082.00490}.

\bibitem[Ding et~al.(2025)Ding, Chen, Huang, Guo, Mao, Shao, Zou, Liu, and Zhang]{ding2025alphaeval}
Hongjun Ding, Binqi Chen, Jinsheng Huang, Taian Guo, Zhengyang Mao, Guoyi Shao, Lutong Zou, Luchen Liu, and Ming Zhang.
\newblock Alphaeval: A comprehensive and efficient evaluation framework for formula alpha mining.
\newblock \emph{arXiv preprint arXiv:2508.13174}, 2025.

\bibitem[Du et~al.(2024{\natexlab{a}})Du, Mao, Xing, and Cambria]{cikm2024_stock_explainable}
Kelvin Du, Rui Mao, Frank Xing, and Erik Cambria.
\newblock Explainable stock price movement prediction using contrastive learning.
\newblock In \emph{Proceedings of the 33rd ACM International Conference on Information and Knowledge Management}, CIKM '24, pp.\  529–537, New York, NY, USA, 2024{\natexlab{a}}. Association for Computing Machinery.
\newblock ISBN 9798400704369.
\newblock \doi{10.1145/3627673.3679544}.
\newblock URL \url{https://doi.org/10.1145/3627673.3679544}.

\bibitem[Du et~al.(2024{\natexlab{b}})Du, Li, Torralba, Tenenbaum, and Mordatch]{10.5555/3692070.3692537}
Yilun Du, Shuang Li, Antonio Torralba, Joshua~B. Tenenbaum, and Igor Mordatch.
\newblock Improving factuality and reasoning in language models through multiagent debate.
\newblock In \emph{Proceedings of the 41st International Conference on Machine Learning}, ICML'24. JMLR.org, 2024{\natexlab{b}}.

\bibitem[Guo et~al.(2024)Guo, Chen, Wang, Chang, Pei, Chawla, Wiest, and Zhang]{ijcai2024p890}
Taicheng Guo, Xiuying Chen, Yaqi Wang, Ruidi Chang, Shichao Pei, Nitesh~V. Chawla, Olaf Wiest, and Xiangliang Zhang.
\newblock Large language model based multi-agents: A survey of progress and challenges.
\newblock In Kate Larson (ed.), \emph{Proceedings of the Thirty-Third International Joint Conference on Artificial Intelligence, {IJCAI-24}}, pp.\  8048--8057. International Joint Conferences on Artificial Intelligence Organization, 8 2024.
\newblock \doi{10.24963/ijcai.2024/890}.
\newblock URL \url{https://doi.org/10.24963/ijcai.2024/890}.
\newblock Survey Track.

\bibitem[Kaplan et~al.(2020)Kaplan, McCandlish, Henighan, Brown, Chess, Child, Gray, Radford, Wu, and Amodei]{scaling_law_openai}
Jared Kaplan, Sam McCandlish, Tom Henighan, Tom~B. Brown, Benjamin Chess, Rewon Child, Scott Gray, Alec Radford, Jeffrey Wu, and Dario Amodei.
\newblock Scaling laws for neural language models, 2020.
\newblock URL \url{https://arxiv.org/abs/2001.08361}.

\bibitem[Ke et~al.(2017)Ke, Meng, Finley, Wang, Chen, Ma, Ye, and Liu]{2017LGBM}
Guolin Ke, Qi~Meng, Thomas Finley, Taifeng Wang, Wei Chen, Weidong Ma, Qiwei Ye, and Tie-Yan Liu.
\newblock Lightgbm: a highly efficient gradient boosting decision tree.
\newblock In \emph{Proceedings of the 31st International Conference on Neural Information Processing Systems}, NIPS'17, pp.\  3149–3157, Red Hook, NY, USA, 2017. Curran Associates Inc.
\newblock ISBN 9781510860964.

\bibitem[Kirkpatrick et~al.(1983)Kirkpatrick, Gelatt, and Vecchi]{kirkpatrick1983optimization}
Scott Kirkpatrick, C~Daniel Gelatt, and Mario~P Vecchi.
\newblock Optimization by simulated annealing.
\newblock \emph{Science}, 220\penalty0 (4598):\penalty0 671--680, 1983.

\bibitem[Koa et~al.(2024)]{SEP2024}
K.~J.~L. Koa et~al.
\newblock Learning to generate explainable stock predictions using self-reflective large language models.
\newblock In \emph{Proceedings of the ACM Web Conference (WWW)}, 2024.
\newblock \doi{10.1145/3589334.3645611}.
\newblock URL \url{https://dl.acm.org/doi/10.1145/3589334.3645611}.

\bibitem[Kovac et~al.(2023)Kovac, Portelas, Dominey, and Oudeyer]{kovac2023the}
Grgur Kovac, R{\'e}my Portelas, Peter~Ford Dominey, and Pierre-Yves Oudeyer.
\newblock The social{AI} school: Insights from developmental psychology towards artificial socio-cultural agents.
\newblock In \emph{First Workshop on Theory of Mind in Communicating Agents}, 2023.
\newblock URL \url{https://openreview.net/forum?id=Y5r8Wa67Ob}.

\bibitem[Li et~al.(2023{\natexlab{a}})Li, Al~Kader~Hammoud, Itani, Khizbullin, and Ghanem]{10.5555/3666122.3668386}
Guohao Li, Hasan~Abed Al~Kader~Hammoud, Hani Itani, Dmitrii Khizbullin, and Bernard Ghanem.
\newblock Camel: communicative agents for "mind" exploration of large language model society.
\newblock In \emph{Proceedings of the 37th International Conference on Neural Information Processing Systems}, NIPS '23, Red Hook, NY, USA, 2023{\natexlab{a}}. Curran Associates Inc.

\bibitem[Li et~al.(2025{\natexlab{a}})Li, Liu, Liu, Fang, Wang, Xu, and Bian]{2025mars}
Junjie Li, Yang Liu, Weiqing Liu, Shikai Fang, Lewen Wang, Chang Xu, and Jiang Bian.
\newblock Mars: a financial market simulation engine powered by generative foundation model.
\newblock In \emph{The Thirteenth International Conference on Learning Representations}, 2025{\natexlab{a}}.
\newblock URL \url{https://openreview.net/forum?id=Yqk7EyT52H}.

\bibitem[Li et~al.(2024{\natexlab{a}})Li, Sun, Lin, Gao, Shang, and Yan]{NEURIPS2024_casualstock}
Shuqi Li, Yuebo Sun, Yuxin Lin, Xin Gao, Shuo Shang, and Rui Yan.
\newblock Causalstock: Deep end-to-end causal discovery for news-driven multi-stock movement prediction.
\newblock In A.~Globerson, L.~Mackey, D.~Belgrave, A.~Fan, U.~Paquet, J.~Tomczak, and C.~Zhang (eds.), \emph{Advances in Neural Information Processing Systems}, volume~37, pp.\  47432--47454. Curran Associates, Inc., 2024{\natexlab{a}}.
\newblock URL \url{https://proceedings.neurips.cc/paper_files/paper/2024/file/54d689d58fe54c92aee2d732fc49fca8-Paper-Conference.pdf}.

\bibitem[Li et~al.(2024{\natexlab{b}})Li, Liu, Shen, Wang, Chen, and Huang]{li2024master}
Tong Li, Zhaoyang Liu, Yanyan Shen, Xue Wang, Haokun Chen, and Sen Huang.
\newblock Master: Market-guided stock transformer for stock price forecasting.
\newblock In \emph{Proceedings of the AAAI Conference on Artificial Intelligence}, volume~38, pp.\  162--170, 2024{\natexlab{b}}.

\bibitem[Li et~al.(2023{\natexlab{b}})Li, Yu, Li, Chen, and Khashanah]{li2023tradinggpt}
Yang Li, Yangyang Yu, Haohang Li, Zhi Chen, and Khaldoun Khashanah.
\newblock Tradinggpt: Multi-agent system with layered memory and distinct characters for enhanced financial trading performance.
\newblock \emph{arXiv preprint arXiv:2309.03736}, 2023{\natexlab{b}}.

\bibitem[Li et~al.(2025{\natexlab{b}})Li, Yang, Yang, Xu, Wang, Liu, and Bian]{rdagentquant}
Yuante Li, Xu~Yang, Xiao Yang, Minrui Xu, Xisen Wang, Weiqing Liu, and Jiang Bian.
\newblock R\&d-agent-quant: A multi-agent framework for data-centric factors and model joint optimization.
\newblock In \emph{The Thirty-ninth Annual Conference on Neural Information Processing Systems}, 2025{\natexlab{b}}.

\bibitem[Liu et~al.(2025)Liu, Guo, Lou, Zeng, Niu, Wang, Xu, Cai, Yang, Zhao, Li, Xu, Chen, Chen, Bai, and Zhang]{liu2025finr1largelanguagemodel}
Zhaowei Liu, Xin Guo, Fangqi Lou, Lingfeng Zeng, Jinyi Niu, Zixuan Wang, Jiajie Xu, Weige Cai, Ziwei Yang, Xueqian Zhao, Chao Li, Sheng Xu, Dezhi Chen, Yun Chen, Zuo Bai, and Liwen Zhang.
\newblock Fin-r1: A large language model for financial reasoning through reinforcement learning, 2025.
\newblock URL \url{https://arxiv.org/abs/2503.16252}.

\bibitem[Liu et~al.(2024)Liu, Zhang, Li, Liu, and Yang]{liu2024dynamicllmpoweredagentnetwork}
Zijun Liu, Yanzhe Zhang, Peng Li, Yang Liu, and Diyi Yang.
\newblock A dynamic llm-powered agent network for task-oriented agent collaboration.
\newblock In \emph{First Conference on Language Modeling}, 2024.

\bibitem[Luo et~al.(2023)Luo, Liao, Li, Cheng, and Yan]{luo-etal-2023-causality}
Di~Luo, Weiheng Liao, Shuqi Li, Xin Cheng, and Rui Yan.
\newblock Causality-guided multi-memory interaction network for multivariate stock price movement prediction.
\newblock In Anna Rogers, Jordan Boyd-Graber, and Naoaki Okazaki (eds.), \emph{Proceedings of the 61st Annual Meeting of the Association for Computational Linguistics (Volume 1: Long Papers)}, pp.\  12164--12176, Toronto, Canada, July 2023. Association for Computational Linguistics.
\newblock \doi{10.18653/v1/2023.acl-long.679}.
\newblock URL \url{https://aclanthology.org/2023.acl-long.679/}.

\bibitem[Miller(1977)]{miller1977risk}
Edward~M Miller.
\newblock Risk, uncertainty, and divergence of opinion.
\newblock \emph{The Journal of Finance}, 32\penalty0 (4):\penalty0 1151--1168, 1977.
\newblock \doi{10.1111/j.1540-6261.1977.tb03317.x}.

\bibitem[Nasdaq(2025)]{Nasdaq100}
Inc. Nasdaq.
\newblock \emph{NASDAQ-100 Index}, 2025.
\newblock URL \url{https://www.nasdaq.com/market-activity/index/ndx}.

\bibitem[Niu et~al.(2022)Niu, Li, and Li]{niu2022metatrader}
Hui Niu, Siyuan Li, and Jian Li.
\newblock Metatrader: An reinforcement learning approach integrating diverse policies for portfolio optimization.
\newblock In \emph{Proceedings of the 31st ACM international conference on information \& knowledge management}, pp.\  1573--1583, 2022.

\bibitem[Park et~al.(2023)Park, O'Brien, Cai, Morris, Liang, and Bernstein]{park2023generative}
Joon~Sung Park, Joseph O'Brien, Carrie~Jun Cai, Meredith~Ringel Morris, Percy Liang, and Michael~S. Bernstein.
\newblock Generative agents: Interactive simulacra of human behavior.
\newblock In \emph{Proceedings of the 36th Annual ACM Symposium on User Interface Software and Technology}, UIST '23, New York, NY, USA, 2023. Association for Computing Machinery.
\newblock ISBN 9798400701320.
\newblock \doi{10.1145/3586183.3606763}.
\newblock URL \url{https://doi.org/10.1145/3586183.3606763}.

\bibitem[Qian et~al.(2025)Qian, Xie, Wang, Liu, Zhu, Xia, Dang, Du, Chen, Yang, Liu, and Sun]{qian2025scaling}
Chen Qian, Zihao Xie, YiFei Wang, Wei Liu, Kunlun Zhu, Hanchen Xia, Yufan Dang, Zhuoyun Du, Weize Chen, Cheng Yang, Zhiyuan Liu, and Maosong Sun.
\newblock Scaling large language model-based multi-agent collaboration.
\newblock In \emph{The Thirteenth International Conference on Learning Representations}, 2025.
\newblock URL \url{https://openreview.net/forum?id=K3n5jPkrU6}.

\bibitem[Qwen et~al.(2025)Qwen, :, Yang, Yang, Zhang, Hui, Zheng, Yu, Li, Liu, Huang, Wei, Lin, Yang, Tu, Zhang, Yang, Yang, Zhou, Lin, Dang, Lu, Bao, Yang, Yu, Li, Xue, Zhang, Zhu, Men, Lin, Li, Tang, Xia, Ren, Ren, Fan, Su, Zhang, Wan, Liu, Cui, Zhang, and Qiu]{qwen2025qwen25technicalreport}
Qwen, :, An~Yang, Baosong Yang, Beichen Zhang, Binyuan Hui, Bo~Zheng, Bowen Yu, Chengyuan Li, Dayiheng Liu, Fei Huang, Haoran Wei, Huan Lin, Jian Yang, Jianhong Tu, Jianwei Zhang, Jianxin Yang, Jiaxi Yang, Jingren Zhou, Junyang Lin, Kai Dang, Keming Lu, Keqin Bao, Kexin Yang, Le~Yu, Mei Li, Mingfeng Xue, Pei Zhang, Qin Zhu, Rui Men, Runji Lin, Tianhao Li, Tianyi Tang, Tingyu Xia, Xingzhang Ren, Xuancheng Ren, Yang Fan, Yang Su, Yichang Zhang, Yu~Wan, Yuqiong Liu, Zeyu Cui, Zhenru Zhang, and Zihan Qiu.
\newblock Qwen2.5 technical report, 2025.
\newblock URL \url{https://arxiv.org/abs/2412.15115}.

\bibitem[Sadka \& Scherbina(2007)Sadka and Scherbina]{sadka2007analyst}
Ronnie Sadka and Anna Scherbina.
\newblock Analyst disagreement, mispricing, and liquidity.
\newblock \emph{The Journal of Finance}, 62\penalty0 (5):\penalty0 2367--2403, 2007.
\newblock \doi{10.1111/j.1540-6261.2007.01278.x}.

\bibitem[Shenzhen Securities Information~Co.(2019)]{ChiNext100_methodology}
Ltd Shenzhen Securities Information~Co.
\newblock \emph{Compilation of the ChixNext 100 Index}, 2019.
\newblock URL \url{https://www.szse.cn/marketServices/message/index/project/P020190201583986359118.pdf}.

\bibitem[Shi et~al.(2025{\natexlab{a}})Shi, Song, Zhang, Shi, Luo, Ao, Arian, and Seco]{AlphaForge}
Hao Shi, Weili Song, Xinting Zhang, Jiahe Shi, Cuicui Luo, Xiang Ao, Hamid Arian, and Luis~Angel Seco.
\newblock Alphaforge: {A} framework to mine and dynamically combine formulaic alpha factors.
\newblock In Toby Walsh, Julie Shah, and Zico Kolter (eds.), \emph{AAAI-25, Sponsored by the Association for the Advancement of Artificial Intelligence, February 25 - March 4, 2025, Philadelphia, PA, {USA}}, pp.\  12524--12532. {AAAI} Press, 2025{\natexlab{a}}.
\newblock \doi{10.1609/AAAI.V39I12.33365}.
\newblock URL \url{https://doi.org/10.1609/aaai.v39i12.33365}.

\bibitem[Shi et~al.(2025{\natexlab{b}})Shi, Duan, and Li]{shi2025navigating}
Yu~Shi, Yitong Duan, and Jian Li.
\newblock Navigating the alpha jungle: An llm-powered mcts framework for formulaic factor mining.
\newblock \emph{arXiv preprint arXiv:2505.11122}, 2025{\natexlab{b}}.

\bibitem[Shi et~al.(2025{\natexlab{c}})Shi, Fu, Chen, Zhao, Xu, Zhang, and Li]{shi2025kronos}
Yu~Shi, Zongliang Fu, Shuo Chen, Bohan Zhao, Wei Xu, Changshui Zhang, and Jian Li.
\newblock Kronos: A foundation model for the language of financial markets.
\newblock \emph{arXiv preprint arXiv:2508.02739}, 2025{\natexlab{c}}.

\bibitem[Tang et~al.(2025)Tang, Chen, Yang, Mai, Zheng, Wang, Chen, and Lin]{tang2025alphaagentllmdrivenalphamining}
Ziyi Tang, Zechuan Chen, Jiarui Yang, Jiayao Mai, Yongsen Zheng, Keze Wang, Jinrui Chen, and Liang Lin.
\newblock Alphaagent: Llm-driven alpha mining with regularized exploration to counteract alpha decay.
\newblock In \emph{Proceedings of the 31st ACM SIGKDD Conference on Knowledge Discovery and Data Mining V. 2}, pp.\  2813--2822, 2025.

\bibitem[Xiao et~al.(2025{\natexlab{a}})Xiao, Sun, Chen, Wu, Luo, and Wang]{xiao2025tradingr1financialtradingllm}
Yijia Xiao, Edward Sun, Tong Chen, Fang Wu, Di~Luo, and Wei Wang.
\newblock Trading-r1: Financial trading with llm reasoning via reinforcement learning, 2025{\natexlab{a}}.
\newblock URL \url{https://arxiv.org/abs/2509.11420}.

\bibitem[Xiao et~al.(2025{\natexlab{b}})Xiao, Sun, Luo, and Wang]{TradingAgents2025}
Yijia Xiao, Edward Sun, Di~Luo, and Wei Wang.
\newblock Tradingagents: Multi-agents {LLM} financial trading framework.
\newblock In \emph{The First MARW: Multi-Agent AI in the Real World Workshop at AAAI 2025}, 2025{\natexlab{b}}.
\newblock URL \url{https://openreview.net/forum?id=4QPrXwMQt1}.

\bibitem[Xu et~al.(2021)Xu, Liu, Wang, Xia, Bian, Yin, and Liu]{xu2021hist}
Wentao Xu, Weiqing Liu, Lewen Wang, Yingce Xia, Jiang Bian, Jian Yin, and Tie-Yan Liu.
\newblock Hist: A graph-based framework for stock trend forecasting via mining concept-oriented shared information.
\newblock \emph{arXiv preprint arXiv:2110.13716}, 2021.

\bibitem[Yang et~al.(2025{\natexlab{a}})Yang, Wang, Jiang, and Wu]{umikdd2025}
Chen Yang, Jingyuan Wang, Xiaohan Jiang, and Junjie Wu.
\newblock Learning universal multi-level market irrationality factors to improve stock return forecasting.
\newblock In \emph{Proceedings of the 31st ACM SIGKDD Conference on Knowledge Discovery and Data Mining V.1}, KDD '25, pp.\  1739–1750, New York, NY, USA, 2025{\natexlab{a}}. Association for Computing Machinery.
\newblock ISBN 9798400712456.
\newblock \doi{10.1145/3690624.3709328}.
\newblock URL \url{https://doi.org/10.1145/3690624.3709328}.

\bibitem[Yang et~al.(2020)Yang, Liu, Zhou, Bian, and Liu]{yang2020qlib}
Xiao Yang, Weiqing Liu, Dong Zhou, Jiang Bian, and Tie-Yan Liu.
\newblock Qlib: An ai-oriented quantitative investment platform.
\newblock \emph{arXiv preprint arXiv:2009.11189}, 2020.

\bibitem[Yang et~al.(2025{\natexlab{b}})Yang, Zhang, Wu, Zhang, Zhang, Yu, Hu, and Wang]{twinmarket}
Yuzhe Yang, Yifei Zhang, Minghao Wu, Kaidi Zhang, Yunmiao Zhang, Honghai Yu, Yan Hu, and Benyou Wang.
\newblock Twinmarket: A scalable behavioral and social simulation for financial markets.
\newblock In \emph{The Thirty-ninth Annual Conference on Neural Information Processing Systems}, 2025{\natexlab{b}}.

\bibitem[Yoo et~al.(2021)Yoo, Soun, Park, and Kang]{2021DTML}
Jaemin Yoo, Yejun Soun, Yong-chan Park, and U~Kang.
\newblock Accurate multivariate stock movement prediction via data-axis transformer with multi-level contexts.
\newblock In \emph{Proceedings of the 27th ACM SIGKDD Conference on Knowledge Discovery \& Data Mining}, KDD '21, pp.\  2037–2045, New York, NY, USA, 2021. Association for Computing Machinery.
\newblock ISBN 9781450383325.
\newblock \doi{10.1145/3447548.3467297}.
\newblock URL \url{https://doi.org/10.1145/3447548.3467297}.

\bibitem[Yu et~al.(2023)Yu, Xue, Ao, Pan, He, Tu, and He]{alphagen}
Shuo Yu, Hongyan Xue, Xiang Ao, Feiyang Pan, Jia He, Dandan Tu, and Qing He.
\newblock Generating synergistic formulaic alpha collections via reinforcement learning.
\newblock In \emph{Proceedings of the 29th ACM SIGKDD Conference on Knowledge Discovery and Data Mining}, 2023.
\newblock \doi{10.1145/3580305.3599831}.

\bibitem[Yu et~al.(2024)Yu, Yao, Li, Deng, Jiang, Cao, Chen, Suchow, Cui, Liu, Xu, Zhang, Subbalakshmi, XIONG, He, Huang, Li, and Xie]{fincon2024}
Yangyang Yu, Zhiyuan Yao, Haohang Li, Zhiyang Deng, Yuechen Jiang, Yupeng Cao, Zhi Chen, Jordan~W. Suchow, Zhenyu Cui, Rong Liu, Zhaozhuo Xu, Denghui Zhang, Koduvayur Subbalakshmi, GUOJUN XIONG, Yueru He, Jimin Huang, Dong Li, and Qianqian Xie.
\newblock Fincon: A synthesized {LLM} multi-agent system with conceptual verbal reinforcement for enhanced financial decision making.
\newblock In \emph{The Thirty-eighth Annual Conference on Neural Information Processing Systems}, 2024.
\newblock URL \url{https://openreview.net/forum?id=dG1HwKMYbC}.

\bibitem[Yuan et~al.(2025)Yuan, Pan, Wang, Gao, Yu, and Yang]{yuanbilevel}
Haochen Yuan, Minting Pan, Yunbo Wang, Siyu Gao, Philip~S. Yu, and Xiaokang Yang.
\newblock Your offline policy is not trustworthy: Bilevel reinforcement learning for sequential portfolio optimization, 2025.
\newblock URL \url{https://arxiv.org/abs/2505.12759}.

\bibitem[Zhang et~al.(2024{\natexlab{a}})Zhang, Liu, Zhang, Jin, Li, Wang, Hua, Shu, Zhu, Jin, Li, Du, and Zhang]{zhang2024aimeetsfinancestockagent}
Chong Zhang, Xinyi Liu, Zhongmou Zhang, Mingyu Jin, Lingyao Li, Zhenting Wang, Wenyue Hua, Dong Shu, Suiyuan Zhu, Xiaobo Jin, Sujian Li, Mengnan Du, and Yongfeng Zhang.
\newblock When ai meets finance (stockagent): Large language model-based stock trading in simulated real-world environments, 2024{\natexlab{a}}.
\newblock URL \url{https://arxiv.org/abs/2407.18957}.

\bibitem[Zhang et~al.(2024{\natexlab{b}})Zhang, Xu, Zhang, Liu, Hooi, and Deng]{zhang-etal-2024-exploring}
Jintian Zhang, Xin Xu, Ningyu Zhang, Ruibo Liu, Bryan Hooi, and Shumin Deng.
\newblock Exploring collaboration mechanisms for {LLM} agents: A social psychology view.
\newblock In Lun-Wei Ku, Andre Martins, and Vivek Srikumar (eds.), \emph{Proceedings of the 62nd Annual Meeting of the Association for Computational Linguistics (Volume 1: Long Papers)}, pp.\  14544--14607, Bangkok, Thailand, August 2024{\natexlab{b}}. Association for Computational Linguistics.
\newblock \doi{10.18653/v1/2024.acl-long.782}.
\newblock URL \url{https://aclanthology.org/2024.acl-long.782/}.

\bibitem[Zhang et~al.(2024{\natexlab{c}})Zhang, Zhao, Xia, Sun, Sun, Qin, Li, Zhao, Zhao, Cai, Zheng, Wang, and An]{FinAgent2024}
Wentao Zhang, Lingxuan Zhao, Haochong Xia, Shuo Sun, Jiaze Sun, Molei Qin, Xinyi Li, Yuqing Zhao, Yilei Zhao, Xinyu Cai, Longtao Zheng, Xinrun Wang, and Bo~An.
\newblock A multimodal foundation agent for financial trading: Tool-augmented, diversified, and generalist.
\newblock In \emph{Proceedings of the 30th ACM SIGKDD Conference on Knowledge Discovery and Data Mining}, KDD '24, pp.\  4314–4325, New York, NY, USA, 2024{\natexlab{c}}. Association for Computing Machinery.
\newblock ISBN 9798400704901.
\newblock \doi{10.1145/3637528.3671801}.
\newblock URL \url{https://doi.org/10.1145/3637528.3671801}.

\bibitem[Zhao et~al.(2024)Zhao, Wang, Zhang, Jin, Zhu, Chen, and Xie]{10.5555/3692070.3694596}
Qinlin Zhao, Jindong Wang, Yixuan Zhang, Yiqiao Jin, Kaijie Zhu, Hao Chen, and Xing Xie.
\newblock Competeai: understanding the competition dynamics of large language model-based agents.
\newblock In \emph{Proceedings of the 41st International Conference on Machine Learning}, ICML'24. JMLR.org, 2024.

\end{thebibliography}
\bibliographystyle{iclr2026_conference}

\appendix
\section{Appendix}
\label{sec:appendix}


\subsection{High-Level workflow of \MYMETHOD}
\label{sec::Appendix::code}
\begin{algorithm}[H]
\caption{\MYMETHOD: Online Learning Framework}\label{alg:searchlayer}
\KwIn{Multi-modal stock features $\mathcal{X}$, macroeconomic data $\mathcal{M}$, historical stock returns $\mathbf{Y}$, number of agent types $n^{\text{type}}$, agents per type $n^{\text{inv}}$, look-back window $\omega_{\text{opt}}$, trading days $T$}
\KwOut{Daily investment portfolio $\mathbf{P}$}

Initialize agent type distribution $\mathbf{d}_{0} \leftarrow [\frac{1}{n^{\text{type}}}, \dots, \frac{1}{n^{\text{type}}}]^\top$\;
\tcp*{Uniform initial distribution}
Initialize all agents $(i,k)$ for $i \in \{1,\dots,n^{\text{type}}\}$, $k \in \{1,\dots,n^{\text{inv}}\}$\;

\For{each trading date $j\in T$}{
  \tcp{--- Forward Propagation: Generate signal for day j ---}
  \For{agent type $i=1$ \KwTo $n^{\text{type}}$}{
    \For{agent $k=1$ \KwTo $n^{\text{inv}}$}{
      Generate investment strategy $\text{Strategy}_{i,j}$ using Eq.~\ref{sec::methods::strgetgy_initalization}\;
      Agent $(i,k)$ selects stocks $\text{Codes}_{i,k,j}$ via Eq.~\ref{sec::methods::investor_decision}\;
    }
  }
  Aggregate agent decisions to compute $\mathrm{Signal}(s,j)$ for all stocks $s$ via Eqs.~\ref{sec::methods::score_agg_1}, \ref{sec::methods::score_agg_2} using distribution $\mathbf{d}_{j-1}$\;
  
  \tcp{--- Portfolio Construction for day j ---}
  Construct portfolio $\mathbf{P}_j$ from $\mathrm{Signal}(:,j)$ using a Top-$k$ strategy\;

  \tcp{--- Backward Optimization: Update distribution for day j+1 ---}
  Optimize distribution $\mathbf{d}_{j}$ using historical data up to day $j$ via Eq.~\ref{sec::methods::distribution_optimization}\;
}
\Return{Sequence of daily portfolios $\{\mathbf{P}_j\}_{j \in T}$}\;
\end{algorithm}

\subsection{Dataset details}
\label{sec:Appendix:dataset}
\subsubsection{Stock pool details}
\begin{itemize}[left=0.2cm, itemsep=-2pt]
    \item \textbf{SSE 50:} This index includes the 50 largest and most liquid stocks on the Shanghai Stock Exchange, mainly large state-owned enterprises and industry leaders. It is stable and blue-chip, suitable for risk-averse and long-term investors focusing on defensive strategies.

    \item \textbf{CSI 300:} Comprising the top 300 stocks from the Shanghai and Shenzhen markets, this index covers diverse industries and company sizes, offering broad market representation. It is ideal for investors seeking diversification and medium- to long-term returns.

    \item \textbf{ChiNext 100:} Featuring 100 stocks from the Shenzhen ChiNext Market, this index focuses on high-tech and innovative firms. Known for its growth potential and higher volatility, it suits investors with high-risk tolerance and those interested in technology sectors.
    \item \textbf{CSI A500:} This index selects 500 leading stocks from A-shares, covering all 35 CSI secondary industries and 91 out of 93 tertiary industries. It emphasizes sector-balanced exposure, ESG screening, and inclusion of innovative "New Quality Productivity" sectors (e.g., IT, industrials, healthcare). With strong profitability (71\% of A-share net profits) and low valuation (14.16x P/E), it serves as a "China S\&P 500" for diversified core-asset allocation and long-term growth strategies. 
\end{itemize}
\subsubsection{Dataset construction details}
We construct our dataset with individual stock data and macroeconomic data.

\textbf{Individual stock data}
\begin{itemize}[left=0.2cm]
    \item \textbf{News}: Stock news is collected from various data sources. We use their titles and summaries as a substitute.
    \item \textbf{Financial Report}: Financial Report is collected from Wind API. We use their titles and summaries as a substitute.
    \item \textbf{E/P\_TTM}: The inverse of the P/E ratio (E/P) indicates the earnings yield, showing the percentage of profit generated per dollar invested in the stock.
    \item \textbf{B/P\_TTM}: Inverse of P/B (B/P) indicates the book yield, showing the return on book value per dollar invested.
    \item \textbf{S/P\_TTM}: The inverse of the price-to-sales ratio (S/P) reflects the sales yield, quantifying the amount of sales revenue generated for each dollar invested in the company. A higher value indicates greater efficiency in converting investment into sales.
    \item \textbf{CF/P\_TTM}: Inverse of P/CF (CF/P) shows the cash flow yield, representing cash flow generated per dollar invested.
    \item \textbf{Log-orthogonalized E/P}: Log-orthogonalized version of E/P, removing some kind of cap basis.Log-orthogonalized version of E/P, removing some kind of cap basis.
    \item \textbf{Log-orthogonalized B/P}:Log-orthogonalized version of the book-to-price ratio, which accounts for and removes certain capitalization effects, thereby isolating the information content of B/P independent of market capitalization.
    \item \textbf{Log-orthogonalized CF/P}:The log-orthogonalized version of the cash flow-to-price ratio, which is employed to control for capitalization influences, ensuring that the ratio captures the true predictive power of cash flow relative to price.
    \item \textbf{Log-orthogonalized S/P}:Log-orthogonalized version of S/P, removing some kind of cap basis.
    \item \textbf{EBITDA/EV}: Measures a company's return on enterprise value, indicating operating earnings (EBITDA) generated per dollar of EV.
    \item \textbf{ROE} : ROE measures profitability by indicating how much net income is generated for each dollar of shareholders’ equity. Higher values signify more effective utilization of equity capital to generate earnings.
    \item \textbf{ROE stability}: TS\_Mean(ROE, 8) / TS\_Std(ROE, 8), measuring both absolute value and stability of ROE.
    \item \textbf{ROA stability}: TS\_Mean(ROA, 8) / TS\_Std(ROE, 8), measuring both absolute value and stability of ROA.
    \item \textbf{Dividend yield}: Dividend yield indicates annual dividends received per dollar invested, expressed as a percentage of the stock price.
    \item \textbf{Log-orthogonalized dividend yield}: Log-orthogonalized version of dividend yield, removing some kind of cap basis.
    \item \textbf{Dividend yield incl repo \& mjrholder trans}: Dividend yield including stock repurchasing and major holder trading.
    \item \textbf{Revenue TTM YoY growth rate}: Measures the percentage change in trailing twelve months' revenue compared to the same period last year.
    \item \textbf{Net profit TTM YoY growth rate}: Measures the percentage change in trailing twelve months' net profit compared to the same period last year.
    \item \textbf{Non-GAAP net profit YoY growth rate}: Indicates the percentage change in non-GAAP net profit compared to the same period last year.
    \item \textbf{Interday volatility}: The price fluctuation range of a stock across trading days.
    \item \textbf{Liquidity}: Weighted average of monthly, quarterly, and yearly turnover ratios.
    \item \textbf{Residual volatility}: Residual volatility measures the unexplained variability in a security's returns after accounting for market or factor influences, indicating idiosyncratic risk.
    \item \textbf{Stock Base data}: The open, high, low, close, volume, and value data of individual stocks on a daily timeframe. (forward-adjusted)
    \item \textbf{industry index return}: One-day return of holding the sector's constituent stocks.
    \item \textbf{Price-volume feature}: Various features extracted from Alpha 158 ~\citep{yang2020qlib} based on price and volume. 
\end{itemize}

\textbf{Macroeconomic data}
\begin{itemize}[left=0.2cm, itemsep=-2pt]
    \item The latest 1-year loan prime rate.
    \item The latest month China CPI YOY growth rate.
    \item The latest yield of China ten ten-year government bonds.
    \item The latest PE and PE quantile of the CSI 300 index.
\end{itemize}

\subsection{More details about \MYMETHOD}
\label{sec::Appendix:moreDetail}

\subsubsection{More details of Market Disagreement hypothesis}
\label{sec::Appendix::marketdis}
Market disagreement describes heterogeneous investor beliefs that drive trading activities. The market disagreement hypothesis posits that such divergence systematically distorts security valuations: when optimistic investors dominate trading while pessimists face short-selling constraints, securities become overpriced and exhibit lower future returns \citep{miller1977risk}. This theory establishes disagreement as a persistent market friction that generates predictable return patterns, with empirical studies confirming that \textbf{high-disagreement stocks consistently underperform consensus-driven counterparts} \citep{diether2002differences, sadka2007analyst}.

\subsubsection{The Design of Investor Initialization}
\label{sec::Appendix:promptInitalization}

\begin{tcolorbox}[
  enhanced,                   
  breakable,                  
  colback=gray!5,             
  colframe=blue!60!black,     
  boxrule=0.8pt,              
  arc=2mm,                    
  left=4pt, right=4pt,        
  top=4pt, bottom=4pt,
  width=\linewidth,           
  title={\bfseries System \& User Prompts},
  fonttitle=\bfseries
]
{\ttfamily\small\justifying

\textbf{System Prompt}

You are a helpful assistant. Make sure you carefully and fully understand the details of the user's requirements before you start solving the problem.

\vspace{1ex}
\textbf{User Prompt}

Give the following input data:

1. Input time-series data column name and their descriptions in JSON format(textual data example).

2. latest macroeconomic and market insights.
Please try to analyze and summarize an abstract investing style description. 

The output format is a json. The specific format of the output JSON is:

\{
   "Outline": "The outline and general description for investment style within 50 words. The outline is a summarization about your investing strategy and your insights into the subsequent trend of the stock market, without any details below.",  
   
   "Details": \{
      "Risk Appetite": "conservative | moderate | moderately conservative | moderately aggressive | aggressive",  
      
      "Holding Period": "one day | about one week | about one month | about half a year | more than one year",  
      
      "Strategy Consistency": [0, 1] (Refers to the investor's ability to adhere to and execute their investment strategy with persistence and coherence, regardless of short-term market fluctuations or emotional influences. Higher number means high consistency",  
      
      "Rationality": [0, 1] (Refers to whether the investor's decision-making process is based on logic, data, and long-term objectives rather than emotions, biases, or short-term market noise. Higher number means high rationality",  
      
      "StockPoolSelector": "Specify what kind of preference you'd like to construct your watchlist stocks. The possible preferences are:
     
      1. RandomStockSelector: Randomly construct your watchlist.
      
      2. IndustryEqualStockSelector: Construct a stock pool with balanced distribution across industries.
      
      3. MVEqualStockSelector: Construct a stock pool with balanced distribution across market capitalizations.
      
      4.IndustryBasisStockSelector: Prefer stocks from specific industries and output the preferred industries. The result is presented in a list format.",
      
      "Others": "Extra information about your investing strategy, maybe correlated with latest market and macroeconmic information and others. No more than 30 words." \} \}

\{examples\}

Input data:

E/P,B/P,CF/P, S/P,Log-orthogonalized E/P,Log-orthogonalized B/P,Log-orthogonalized CF/P,Log-orthogonalized S/P,

Macro data:

The latest 1-year loan prime rate is 3.45. The latest month China CPI YOY growth rate is -0.5. The latest yield of China's ten-year government bonds is 2.6733\%, while the yield has increased 0 BP over the past one day, increased -4 BP over the past one month, and increased -21 BP over the past half a year. The latest CSI\_300 PE is 10.9478, and the current PE ratio of the CSI 300 is at the 5.4th percentile over the past 5 years(0 indicates most undervalued, and 100 indicates most overvalued). The latest market sentiment index got a 0.63\% return.

Your investing style:

\{'Outline': 'A value-oriented investment approach focusing on fundamentally strong companies with a long-term perspective, leveraging current market undervaluation and stable economic indicators to build a diversified portfolio.',

 'Details': \{'Risk Appetite': 'moderate',
  'Holding Period': 'more than one year',
  'Strategy Consistency': '0.85',
  'Rationality': '0.9',
  'StockPoolSelector': 'IndustryEqualStockSelector',
  'Others': 'Leverage low CPI and undervalued CSI 300 PE for potential upside.'\}
  
(END\_OF\_EXAMPLES)

Input data:
\{input\_data\}

Macro data:
\{macro\_data\}

Your investing style:

} 
\end{tcolorbox}

\subsubsection{The design of Investment Strategy execution}
\label{sec::Appendix:promptExecution}

\begin{tcolorbox}[
  enhanced,                   
  breakable,                  
  colback=gray!5,             
  colframe=blue!60!black,     
  boxrule=0.8pt,              
  arc=2mm,                    
  left=4pt, right=4pt,        
  top=4pt, bottom=4pt,
  width=\linewidth,           
  title={\bfseries User Prompts},
  fonttitle=\bfseries
]
{\ttfamily\small\justifying

\textbf{User Prompt}

Giving following 

1. Input data in table format and their descriptions in JSON format.

2. investing style to make investment decisions in JSON format.

Please output \{num\_stocks\} stocks you tend to invest in. The result is in JSON format, key is "Stock", and value is a list containing the stock code. Please make sure:

1. You output legal stock code. The stock code is legal if and only if it is in the input data "Stock" list.

2. The number of stock codes is correct, actually equal to \{num\_stocks\}.
Here is an example.

For stock\_nums in investing instructions, we use 3 in this example.
Input Data for investing decision:

1. \textbf{Input Data Description}:

\{"E/P": "The inverse of the P/E ratio (E/P) indicates the earnings yield, showing the percentage of profit generated per dollar invested in the stock.",

"B/P": "Inverse of P/B (B/P) indicates the book yield, showing the return on book value per dollar invested.",

"S/P": "Inverse of P/S (S/P) reflects the sales yield, showing sales generated per dollar invested.",

"CF/P": "Inverse of P/CF (CF/P) shows the cash flow yield, representing cash flow generated per dollar invested.",

"Log-orthogonalized E/P": "Log-orthogonalized version of E/P, removing some kind of cap basis.",

"Log-orthogonalized B/P": "Log-orthogonalized version of B/P, removing some kind of cap basis.",

"Log-orthogonalized CF/P": "Log-orthogonalized version of CF/P, removing some kind of cap basis.",

"Log-orthogonalized S/P": "Log-orthogonalized version of S/P, removing some kind of cap basis.",

"EBITDA/EV": "Measures a company's return on enterprise value, indicating operating earnings (EBITDA) generated per dollar of EV."\}

2. \textbf{Investing Style}:

\{"Outline": "A value-driven investment approach focusing on stocks with strong fundamentals, undervalued valuations, and consistent cash flows over the long term.",  
   
   "Details": \{
      "Risk Appetite": "Moderately conservative",  
      "Holding Period": "More than one year",  
      "Strategy Consistency": "0.85",  
      "Rationality": "0.9",  
      "StockPoolSelector": "MVEqualStockSelector"  \}\}

3. \textbf{Input data}:

',Stock,Date,E/P,B/P,CF/P,S/P,

Log-orthogonalized E/P,Log-orthogonalized B/P,Log-orthogonalized CF/P,

Log-orthogonalized S/P, EBITDA/EV,

965494,000858,20190102, 0.06295366,

0.30744636,0.038947526,0.19324197,

-4.032941,-1.1295723,3.594055,

-1.2754831,0.124886042941460,

002594,20190102,0.020888906,

0.37708813,0.09185906,0.9017491,-4.038043,

-0.6966869,5.084233,0.3152281,0.09258402716,

600519,20190102,0.042301364,0.13605072,

0.036664255,0.09038502,-7.6968794,-2.2439895,

1.2049837,-2.2207088,0.0797575348104294,

600900,20190102,0.066111766,0.4052357,0.1183,

0.15322393,-5.3881683,-1.0025798,3.743841,

-1.5840118,0.1050353948267292,601012,

20190102,0.062190603,0.30756927,0.032795224,

0.41643697,-0.72993636,-0.7708632,

5.801872,-0.31826368,0.0887390158431868,

601288,20190102,0.16604953,1.2584949,

0.12149128,0.4757528,-7.5973797,-0.1158539,

1.556502,-0.6272717,0.059454748665067,

601888,20190102,0.02850359,0.1358013,

0.034710173,0.35662726,-3.433404,-1.7193639,

4.34933,-0.59344673,0.0511954139068489,

603259,20190102,0.024971908,0.12955885,

0.018961666,0.10751114,-2.9358995,

-1.8100101,4.314365,-1.7471998,0.04303389'

\textbf{LLM output}:

\{'Stock': ['000858', '600900', '601288']\}

Note that in this example, we ask LLM to output 3 stocks. However, in real scenarios, you should follow the "num\_stocks" args in the instruction.

(END OF EXAMPLES)

\{input\_data\}

} 
\end{tcolorbox}

\setlength{\tabcolsep}{4.5pt}

\subsubsection{Time complexity and token cost}
\label{sec::Appendix:moreDetail:timeandcost}


\begin{table}[h!]
\centering
\caption{\MYMETHOD~'s average time cost and api call fees on each trading day.}
\label{tab:Cost}
\begin{tabular}{@{}lrr@{}}
\toprule
\textbf{Stock Pool} & \textbf{Time} & \textbf{Cost} \\
\midrule
SSE50 & 125s & \$0.679 \\
CSI 300 & 378s & \$2.265 \\
Chi Next 100 & 227s & \$1.192 \\
\bottomrule
\end{tabular}
\end{table}
For practical deployment, the operational efficiency of \MYMETHOD~ is a key consideration. We have incorporated several strategic design choices to minimize API-related costs and ensure computational feasibility in a live trading scenario.

First, we recognize that the macroeconomic data informing the investor initialization module (Section~\ref{sec::method::investor_initialization}) changes at a low frequency. For instance, metrics like the CPI growth rate are updated monthly, while others, such as government bond yields and market-wide PE percentiles, evolve gradually. To substantially mitigate API overhead, we therefore limit the re-invocation of this initialization module to only once per week, specifically on the first trading day.

Second, and critically, we cache the individual investment decisions generated by each agent during the forward propagation stage. Consequently, the backward optimization process—while computationally intensive in its re-aggregation of historical signals for different candidate distributions—operates entirely on these cached results. This design ensures that the backward optimization step incurs \textbf{zero additional LLM token costs}.

These optimization strategies ensure the economic viability of \MYMETHOD~ in a real-world setting. To provide a transparent assessment of its practicality, we report the average daily computational time and API costs for a 512-agent configuration in Table~\ref{tab:Cost}.
\subsection{Baseline details}
\label{sec:Appendix:baselines}

\begin{itemize}[left=0.2cm, itemsep=-2pt]
    \item  \textbf{Proxy indicators}: Various features can be used as a proxy to quantify market disagreement \citep{diether2002differences}. We use the earning stability of the listed company (implemented by calculating the std of annualized ROE) as a baseline.
    \item \textbf{LightGBM}: A high-efficiency, leaf-wise gradient boosting decision tree framework by Microsoft Research, employing histogram-based algorithms for accelerated training and reduced memory footprint. Following \citep{NBER2023} and Equation \ref{sec::methods::score_agg_2}, we simulate market disagreement by constructing various LightGBM agents visible to different features.
    \item \textbf{DTML}: DTML ~\citep{2021DTML} is an attention-based model that exploits the correlations between stocks to make investment decisions. We use the open-source implementation \footnote{\url{https://github.com/ceteris11/DTML}} to implement this baseline.
    \item \textbf{MASTER}: MASTER ~\citep{li2024master} is a stock transformer for stock price forecasting, which models the momentary and cross-time stock correlation and guides feature selection with market information. We use the open-source implementation \footnote{\url{https://github.com/SJTU-DMTai/MASTER}} to implement this baseline.
    \item \textbf{SEP}: SEP ~\citep{SEP2024} utilizes a verbal self-reflective agent and
    A PPO that allows the LLM to teach itself how to generate explainable single stock predictions. We use the open-source link \footnote{\url{https://github.com/koa-fin/sep}} to implement SEP.
    \item \textbf{FINCON}: FINCON ~\citep{fincon2024} is a multi-agent framework for single stock price prediction and simple investment portfolio construction with conceptual verbal reinforcement.
    \item \textbf{TradingAgents}: TradingAgents ~\citep{TradingAgents2025} is a multi-agent framework.
    that utilizes trading firms' collaborative dynamics to construct investment portfolios.  We use the open-source link \footnote{\url{https://github.com/TauricResearch/TradingAgents}} to implement TradingAgents.
\end{itemize}

\begin{table*}[ht!]
\centering
\caption{Comparisons with baselines on more evaluation metrics. \MYMETHOD~outperforms all others across all 3 stock pools, showing impressive cumulative returns compared to the stock index. The best performance in each column is highlighted in \textbf{bold}.}
\label{tab:MoreEvaluationMetrics}
\scalebox{0.9}{
\begin{tabular}{@{}lrrrrrrrrr@{}}
\toprule
\multicolumn{10}{c}{\textbf{Main Experiments (Throughout 2023)}} \\ \midrule
\multicolumn{1}{c}{\multirow{2}{*}{\textbf{Method}}} & \multicolumn{3}{c}{\textbf{SSE50}} & \multicolumn{3}{c}{\textbf{CSI 300}} & \multicolumn{3}{c}{\textbf{Chi Next 100}} \\ 
\cmidrule(lr){2-4} \cmidrule(lr){5-7} \cmidrule(lr){8-10}
\multicolumn{1}{c}{} & \multicolumn{1}{c}{AR} & \multicolumn{1}{c}{Sharpe} & \multicolumn{1}{c}{MDD} & \multicolumn{1}{c}{AR} & \multicolumn{1}{c}{Sharpe} & \multicolumn{1}{c}{MDD} & \multicolumn{1}{c}{AR} & \multicolumn{1}{c}{Sharpe} & \multicolumn{1}{c}{MDD} \\ \midrule
Proxy Indicator~\citep{diether2002differences} & -2.39 & -1.22 & 14.04 & -3.60 & -1.62 & 20.57 & -20.01 & -3.24 & 24.15 \\
LightGBM ~\citep{2017LGBM} & -1.88 & -1.14 & 13.16 & -4.55 & -2.12 & 18.57 & -19.32 & -3.01 & 23.96 \\
DTML~\citep{2021DTML} & -1.69 & -1.08 & 12.99 & -0.33 & -0.14 & 22.34 & -8.23 & -3.20 & 24.55 \\
MASTER~\citep{li2024master} & -1.67 & -0.92 & 12.91 & 0.79 & 0.33 & 22.05 & -7.88 & -3.17 & 24.06 \\
SEP~\citep{SEP2024} & -2.01 & -1.07 & 13.12 & -10.24 & -4.32 & 22.67 & -6.84 & -3.14 & 24.01 \\
FinCON~\citep{fincon2024} & -1.82 & -0.98 & 13.05 & -9.25 & -3.28 & 23.74 & -6.01 & -2.80 & 23.75 \\
TradingAgents~\citep{TradingAgents2025} & -2.44 & -1.71 & 13.15 & -7.19 & -3.02 & 19.61 & -4.65 & -2.82 & 23.84 \\ 
\rowcolor{gray!20}
\textbf{\MYMETHOD} & \textbf{2.16} & \textbf{1.98} & \textbf{11.98} & \textbf{4.95} & \textbf{2.23} & \textbf{14.04} & \textbf{1.17} & \textbf{0.99} & \textbf{19.06} \\ \midrule
Stock pool Index & -9.98 & -2.37 & 21.62 & -9.75 & -2.92 & 21.44 & -19.18 & -3.17 & 32.26 \\ \midrule
\multicolumn{10}{c}{\textbf{Experiments on data leakage concern (The first quarter of 2025)}} \\ \midrule
\multicolumn{1}{c}{\multirow{2}{*}{\textbf{Method}}} & \multicolumn{3}{c}{\textbf{SSE50}} & \multicolumn{3}{c}{\textbf{CSI 300}} & \multicolumn{3}{c}{\textbf{CSI A500}} \\ 
\cmidrule(lr){2-4} \cmidrule(lr){5-7} \cmidrule(lr){8-10}
\multicolumn{1}{c}{} & \multicolumn{1}{c}{AR} & \multicolumn{1}{c}{Sharpe} & \multicolumn{1}{c}{MDD} & \multicolumn{1}{c}{AR} & \multicolumn{1}{c}{Sharpe} & \multicolumn{1}{c}{MDD} & \multicolumn{1}{c}{AR} & \multicolumn{1}{c}{Sharpe} & \multicolumn{1}{c}{MDD} \\ \midrule
\rowcolor{gray!20}
\textbf{\MYMETHOD} & \textbf{9.74} & \textbf{2.42} & \textbf{2.91} & \textbf{9.36} & \textbf{2.66} & \textbf{2.99} & \textbf{11.34} & \textbf{2.93} & \textbf{4.08} \\ 
Stock pool Index & -1.88 & -2.97 & 5.63 & -3.88 & -3.15 & 5.86 & -1.28 & -3.26 & 6.04 \\ \bottomrule
\end{tabular}
}
\end{table*}

\subsection{Further experiment results on more evaluation metrics}

\label{sec::Appendix::furtherRes}
In this section, we provide more evaluation metrics to demonstrate the superiority of \MYMETHOD. These metrics are as follows: 

\begin{enumerate}
    \item \textbf{Annualized Return (AR)} \\
    The average annual return of the strategy, calculated by scaling the periodic return (e.g., daily, monthly) to an annual basis. It reflects the strategy's profitability over time, with the formula:
    \[
    R_{\text{annual}} = \left(1 + R_{\text{periodic}}\right)^{n} - 1
    \]
    where $n$ is the number of periods in a year.

    \item \textbf{Maximum Drawdown (MDD)} \\
    The largest peak-to-trough decline in portfolio value, expressed as a percentage. It measures the strategy's downside risk, defined as:
    \[
    \text{MDD} = \max\left(1 - \frac{V_t}{V_{\text{peak}}}\right)
    \]
    where $V_t$ is the portfolio value at time $t$, and $V_{\text{peak}}$ is the maximum value before $t$.

    \item \textbf{Sharpe Ratio (Sharpe)} \\
    A measure of risk-adjusted return, calculated as the excess return over the risk-free rate divided by the strategy's volatility:
    \[
    \text{SR} = \frac{R_{\text{strategy}} - R_f}{\sigma_{\text{strategy}}}
    \]
where $R_f$ is the risk-free rate and $\sigma_{\text{strategy}}$ denotes the standard deviation of strategy returns. A higher value indicates superior risk-adjusted performance.
\end{enumerate}

\subsection{Backtesting experiments details}
\label{sec::Appendix::backtesting}
We conduct backtesting experiments on a simulated system. We conduct backtesting using a traditional index-enhancement strategy. The portfolio is rebalanced weekly, with a round-trip transaction cost of $0.1\%$. During the first fifteen minutes after the market opens on the first trading day of each week, we first exclude stocks that are limit-up or limit-down. Subsequently, we rank the portfolio construction signals and equally weight the top 20\% of the ranked stocks. Stocks currently held but no longer in the top 20\% are sold.

\end{document}